\documentclass[sigconf]{acmart}

\copyrightyear{2024}
\acmYear{2024}
\setcopyright{acmlicensed}
\acmConference[MM '24] {Proceedings of the 32nd ACM International Conference on Multimedia}{October 28--November 1, 2024}{Melbourne, VIC, Australia.}
\acmBooktitle{Proceedings of the 32nd ACM International Conference on Multimedia (MM '24), October 28--November 1, 2024, Melbourne, VIC, Australia}
\acmISBN{979-8-4007-0686-8/24/10}
\acmDOI{10.1145/3664647.3680827}

\begin{document}

\title{FineCLIPER: Multi-modal Fine-grained CLIP for Dynamic Facial Expression Recognition with AdaptERs}


\author{Haodong Chen}
\orcid{0009-0009-1666-2037}
\affiliation{%
  \institution{School of Automation, Northwestern Polytechnical University}
  \city{Xi'an City}
  \country{China}}
\email{chd@mail.nwpu.edu.cn}

\author{Haojian Huang}
\orcid{0000-0002-0661-712X}
\affiliation{%
  \institution{The University of Hong Kong}
  \city{Hong Kong}
  \country{China}}
\email{haojianhuang927@gmail.com}

\author{Junhao Dong}
\orcid{0000-0002-6232-9157}
\affiliation{%
  \institution{Nanyang Technological University}
  \country{Singapore}}
\email{junhao003@ntu.edu.sg}

\author{Mingzhe Zheng}
\orcid{0009-0000-9721-2592}
\affiliation{%
  \institution{School of Computer Science, Northwestern Polytechnical University}
  \city{Xi'an City}
  \country{China}}
\email{normanzheng6606@gmail.com}

\author{Dian Shao}
\orcid{0000-0002-0862-9941}
\affiliation{%
  \institution{Unmanned System Research Institute, Northwestern Polytechnical University}
  \city{Xi'an City}
  \country{China}}
\email{shaodian@nwpu.edu.cn}
\authornote{Corresponding author}

\renewcommand{\shortauthors}{Haodong Chen, Haojian Huang, Junhao Dong, Mingzhe Zheng, and Dian Shao}

\begin{abstract}
Dynamic Facial Expression Recognition (DFER) is crucial for understanding human behavior.
However, current methods exhibit limited performance mainly due to the insufficient utilization of facial dynamics, and the ambiguity of expression semantics, etc.
To this end, we propose a novel framework, named Multi-modal \underline{Fine}-grained \underline{CLIP} for DFER with Adapt\underline{ER}s (\textbf{FineCLIPER}), incorporating the following novel designs: 
1) To better distinguish between similar facial expressions, we extend the class labels to textual descriptions from both positive and negative aspects, and obtain supervision by calculating the cross-modal similarity based on the CLIP model;
2) Our FineCLIPER adopts a hierarchical manner to effectively mine useful cues from DFE videos. Specifically, besides directly embedding video frames as input (\textit{low semantic level}),
we propose to extract the face segmentation masks and landmarks based on each frame (\textit{middle semantic level}) and utilize the Multi-modal Large Language Model (MLLM) to further generate detailed descriptions of facial changes across frames with designed prompts (\textit{high semantic level}).
Additionally, we also adopt Parameter-Efficient Fine-Tuning (PEFT) to enable efficient adaptation of large pre-trained models (\textit{i.e.}, CLIP) for this task.
Our FineCLIPER achieves SOTA performance on the DFEW, FERV39k, and MAFW datasets in both supervised and zero-shot settings with few tunable parameters. 
Project page: \textcolor{magenta}{\url{https://haroldchen19.github.io/FineCLIPER-Page/}}

\end{abstract}





\maketitle

\begin{figure}[!t]
    \centering
    \vspace{-0.4em}
   \includegraphics[width=1\linewidth]{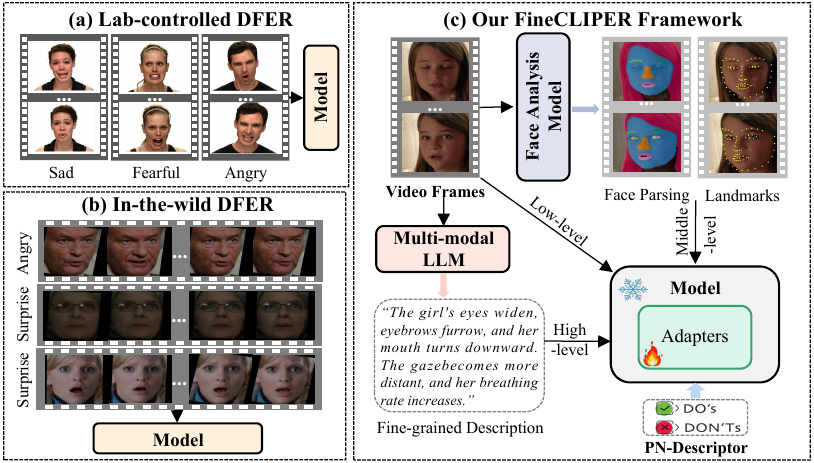}
   \vspace{-2em}
    \captionof{figure}{\label{fig:dataset}
    Frameworks for DFER.}
    \vspace{-6mm}
\end{figure}

\vspace{-0.4em}
\section{Introduction}
Facial expressions are important signals to convey human emotions,
thus accurately recognizing them has significant meaning for various tasks, including interpersonal communication, human-computer interaction (HCI)~\cite{8039024, ladak2024artificial, hundt2024love}, mental health diagnosing~\cite{kumar2024measuring, sarkar2023towards, 9674818}, driving safety monitoring~\cite{wang2023driver, yan2024empower, zhang2024two}, etc. 
Traditional Facial Expression Recognition (FER) resorts to static images. 
However, since dynamic emotional changes could not be well-represented within a single image,
research attention has been shifted to Dynamic Facial Expression Recognition (DFER), which distinguishes the temporally displayed facial expressions in videos.

The study of DFER algorithms starts from highly-controlled environments~\cite{6849440, 10.1371/journal.pone.0196391, 1623803}, where faces are frontal and non-blurry~\cite{Li_2022, wang2022systematic} as shown in Fig.~\ref{fig:dataset} (a). 
However, such an ideal assumption makes the obtained models vulnerable to real-world situations. 
Therefore, researchers have turned to more open scenes and constructed several \textit{in-the-wild} DFER datasets,
\textit{e.g.}, DFEW~\cite{jiang2020dfew}, FERV39k \cite{wang2022ferv39k}, and MAFW~\cite{liu2022mafw},
to facilitate the development of corresponding methods~\cite{zhao2021former, wang2023rethinking, li2023intensity, ma2022spatio, li2022nr, liu2023expression} as demonstrated in Fig.~\ref{fig:dataset} (b).
While category labels are treated without semantic meanings (\textit{e.g.}, \textit{Happiness} may only be represented by a class id \textit{"0"}),
recent research on DFER~\cite{zhao2023prompting, foteinopoulou_emoclip_2024,tao2024a3ligndferpioneeringcomprehensivedynamic,li2023cliper} has further delved into the exploration of vision-language multi-modal learning beyond the traditional classification paradigms~\cite{kossaifi2020factorized, sun2020multi, baddar2019mode, 10.1145/3503161.3547865, li2023intensity, 10.1145/3474085.3475292}.

Although huge efforts have been spent, the performance of DFER methods still suffers from noisy frames, small inter-class differences, and ambiguity between expressions, making it inappropriate to adopt video/action recognition techniques directly.
Specifically, to distinctly improve the performance of DEFR algorithms, we have to face the following unique and tough challenges:
1) The ambiguity of semantic labels for dynamic facial expressions, and
2) The subtle and nuanced movements of local face parts (\textit{i.e.}, skeletons, muscles, etc.).
The first challenge originates in the difficulty of accurate human labeling and complex expression ways adopted by different persons, while the latter demands additional focus on fine-grained details that happen in specific regions within a human face.

\begin{figure*}
    \centering
    \vspace{-0.6em}
   \includegraphics[width=1\linewidth]{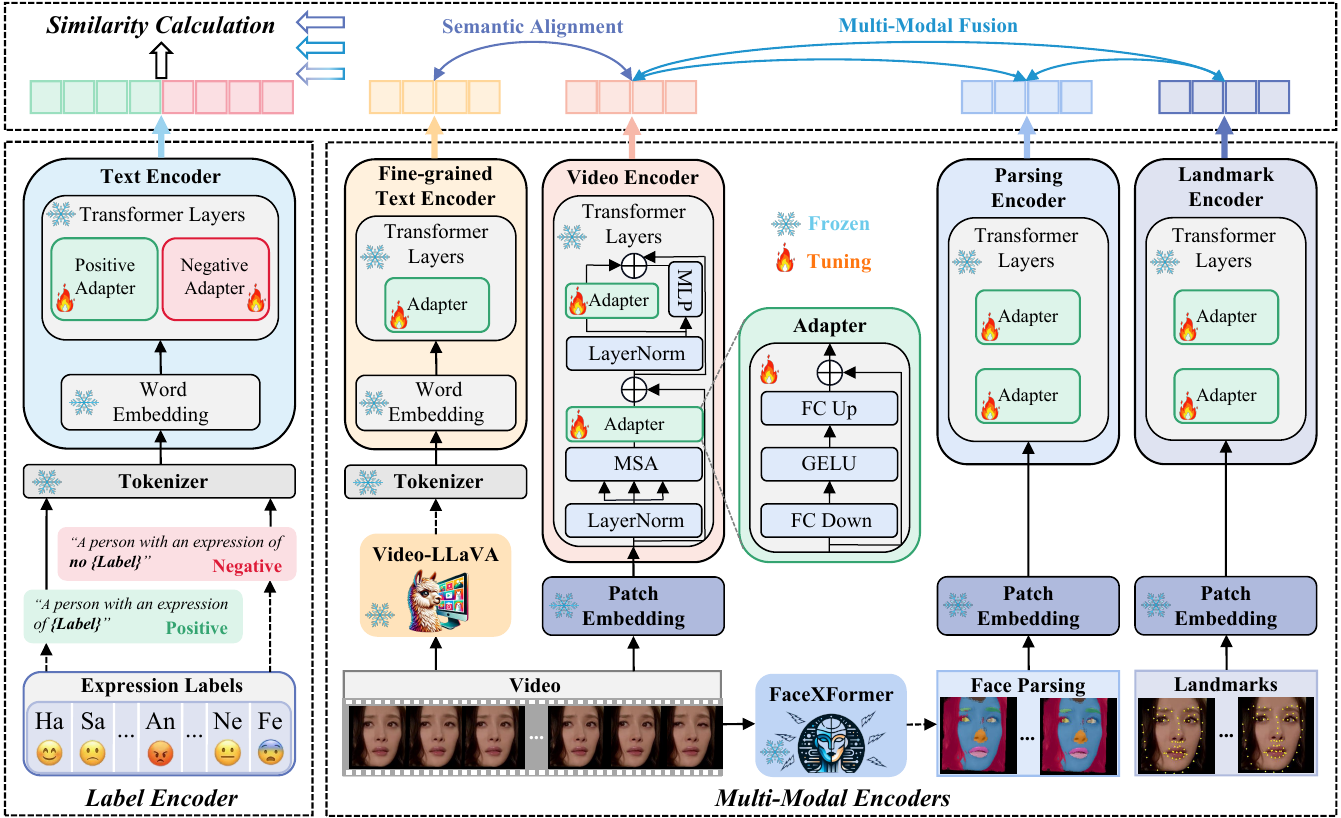}
   \vspace{-2em}
    \captionof{figure}{\label{fig:framework}
    The FineCLIPER framework can be divided into three main components: Label Encoder, Multi-Modal Encoders, and Similarity Calculation. The Label Encoder augments labels using PN descriptors, followed by PN adaptors within text encoder; The Multi-Modal Encoders handle hierarchical information mined from low semantic levels to high semantic levels of human face; The Similarity Calculation module further integrates and computes the similarities of the representations obtained earlier via contrastive learning.}
    \vspace{-0.8em}
\end{figure*}

To tackle the above challenges, we propose a novel framework called FineCLIPER, short for Multi-modal \underline{Fine}-grained \underline{CLIP} for Dynamic Facial Expression Recognition with Adapt\underline{ER}s. 
Specifically, we utilize the Contrastive Language-Image Pretraining (CLIP) model~\cite{radford2021learning}, which is
particularly suitable for providing a cross-modal latent space. 
To avoid the huge cost of fine-tuning such large pre-trained models, FineCLIPER adopts the Parameter-Efficient Fine-Tuning (PEFT) strategy by adding several adaption modules with small parameters for tuning (as shown in Fig.~\ref{fig:framework}), achieving high efficiency while preserving the remarkable performance. 

Specifically, our FineCLIPER has the following characteristics that distinguish it from previous works:

Firstly, by adopting the vision-text learning paradigm, we transform the ground truth label to form the textual supervision (\textit{e.g.}, \textit{"A person with an expression of \{Label\}"}). But one noteworthy innovation is that we meanwhile generate and use the negative counterparts (\textit{e.g.}, \textit{"A person with an expression of \textbf{No} \{Label\}"}).
Such label augmentation via PN (Positive-Negative) descriptors is inspired by the negative prompting strategy ~\cite{miyake2023negative, dong2022dreamartist}, 
and found to be useful here for differentiating between ambiguous categories.
A notable progress is observed in the "Disgust" category of the DFEW dataset, while most baselines~\cite{liu2022clip, li2022nr, li2023intensity, wang2023rethinking, chen2023static} suffer from a nearby  0\% accuracy, our FineCLIPER significantly promotes the performance by more than 25\%, as shown in Tab.~\ref{tab:specific}. 

Furthermore, we adopt a \textit{semantically hierarchical strategy} to comprehensively mine useful information from the input video data.
Specifically, features from directly embedding video frames stand at a relatively \textit{low} semantic level.
For \textit{middle} semantic level, we utilize a well-trained face analysis model (\textit{i.e.} FaceXFormer~\cite{narayan2024facexformer}) to extract the face segmentation masks and landmarks from each frame. Intuitively, the former offers prior about face structures
while the latter provides specific pivots for model attention.
Additionally, we try to obtain descriptions at a \textit{high} semantic level for describing dynamic facial changes across frames. This is realized by leveraging a well-trained MLLM, Video-LLaVA~\cite{lin2023video} to act as a facial expression analyst following given template-based prompts,
and the generated descriptions will be carefully refined.
All the above features at various semantic levels will be integrated to obtain the final representation of a given video. To summarize, our contributions are as follows:

\begin{itemize}[leftmargin=*]
    \item We introduce FineCLIPER, a novel multi-modal framework that enhances Dynamic Facial Expression Recognition (DFER) through extensively mining useful information at different semantic levels from the video data, and all the obtained features (\textit{i.e.}, features embedded from visual frames, face segmentation, face landmarks, and the extra fine-grained descriptions obtained via MLLM are integrated to serve as a more comprehensive representation;
    \item To address the ambiguity between categories, we propose a label augmentation strategy, not only transforming the class label to textual supervision but also using a combination of both positive and negative descriptors;
    \item Extensive experiments conducted on DFER datasets, \textit{i.e.}, DFEW, FERV39k, and MAFW, show that our FineCLIPER framework achieves new state-of-the-art performance on both supervised and zero-shot settings with only a small number of tunable parameters. Comprehensive ablations and analyses further validate the effectiveness of FineCLIPER.
    \vspace{-0.5em}
\end{itemize}

\section{Related Work}
\noindent{\bf Dynamic Facial Expression Recognition.}
In early DFER research, the focus was on developing diverse local descriptors on lab-controlled datasets~\cite{6849440, Livingstone2018TheRA, 10.1371/journal.pone.0196391}. 
Then the rise of deep learning and accessible in-the-wild DFER datasets~\cite{jiang2020dfew, wang2022ferv39k, liu2022mafw} leads to new trends towards DFER research.
The first trend~\cite{fan2016video, kossaifi2020factorized, lee2019context} involves the direct use of 3D CNNs~\cite{tran2015learning, hara2018can, tran2018closer} to extract joint spatio-temporal features from raw videos. The second trend~\cite{ebrahimi2015recurrent, kollias2020exploiting, sun2020multi, 10.1145/3503161.3547865} combines 2D CNNs~\cite{simonyan2014very, chung2014empirical} with RNNs~\cite{chung2014empirical, hochreiter1997long} for feature extraction and sequence modeling. The third emerging trend~\cite{10.1145/3474085.3475292, ma2022spatio, li2023intensity} integrates transformer~\cite{dosovitskiy2020image}. These methods combine convolutional and attention-based approaches to enhance the understanding of visual data, especially in distinguishing samples based on varying visual dynamics.
However, in prior efforts, 
the semantic meaning of class labels is neglected, and insufficient attention has been paid to the subtle and nuanced movements of the human face.
Therefore, based on the well-trained large cross-modal models (\textit{i.e.}, CLIP), we propose to extend the class label to textual supervision both positively and negatively. Moreover, to fully exploit the visual information within videos, we also design a hierarchical information mining strategy to generate representative video features, which is a weighted fusion of various features involving different semantic levels, including video frame feature, the middle-level facial semantics from segmentation maps and detected landmarks, as well as the high-level semantics encoded from fine-grained descriptions.

\noindent{\bf CLIP in Classification.} With the development of computer vision~\cite{chen2024gaussianvton, huang2024crest, shao2020finegym, shao2018find, shao2020intra, wang2023prototype}, Vision-Language Models~\cite{jiang2024effectiveness}, \textit{e.g.}, CLIP~\cite{radford2021learning}, have recently demonstrated superior performance across various tasks~\cite{wang2024m2, yu2024tf, yan2024urbanclip}. 
Recent studies~\cite{zhao2023prompting, tao2024a3ligndferpioneeringcomprehensivedynamic,li2023cliper} have also applied CLIP to the DFER task. 
Among them, A$^3$lign-DFER~\cite{tao2024a3ligndferpioneeringcomprehensivedynamic} introduces a comprehensive alignment paradigm for DFER through a complicated design. CLIPER~\cite{li2023cliper} adopts a two-stage training paradigm instead of end-to-end training; however, it is limited in capturing temporal information. Furthermore, DFER-CLIP~\cite{zhao2023prompting} incorporates a transformer-based module to better capture temporal information in videos, but it requires fully fine-tune the image encoder and the proposed temporal module during training, leading to inefficiency. 

However, while these works have explored the semantic information of labels compared to traditional DFER, they often overlook the interrelations among facial expressions and the individual differences among humans as they directly extend labels into relevant action descriptions (\textit{e.g.}, Happiness$\rightarrow$smiling mouth, raised cheeks, wrinkled eyes, ...~\cite{zhao2023prompting}). This oversight can lead to further ambiguity. 
In light of this, we propose PN (Positive-Negative) descriptors,
extending the ground truth labels from contrastive views to better distinguish between ambiguous categories.

\vspace{-0.4em}
\section{Methodology}

This section outlines the framework's overall pipeline and basic notations (Sec.~\ref{pipeline}), the augmentation of class labels for positive-negative textual supervision (Sec.~\ref{labelencoder}), the hierarchical information mining strategy (Sec.~\ref{mining}), and the integration of these features (Sec.~\ref{fusion}). The complete pipeline is shown in Fig.~\ref{fig:framework}.
\vspace{-1em}

\subsection{Overall Pipeline}
\label{pipeline}
Formally, given a video clip $V$,
the task of DFER aims to recognize the facial expression label $Cls$.
Using text templates as \textit{"A person with an expression of \{Cls\}"}, the class label could be further transformed into textual supervision, which could better utilize the semantic meaning of the category name.

Let $\mathcal{V}$ represents a set of videos and $\mathcal{C}$ denotes collections of augmented textual descriptions of labels, our framework could produce the embedded representations for both a given video and its corresponding textual supervision, resulting in $\mathbf{v_i}$ and $\mathbf{c_i}$.
Note that in our cases,  $\mathbf{v_i}$ is an integration of features from different semantic levels, namely low-level (video frames), middle-level (face parsing and landmarks), and high-level semantics (fine-grained captions of facial action changes obtained using MLLM).
The similarity between $\mathbf{v_i}$ and $\mathbf{c_i}$ is calculated as $sim_i$.
To employ the cross-entropy loss, we calculate the prediction probability over class $cls_i$ as:
\begin{equation}
    p(cls_i|\mathbf{v_i})=\frac{\exp(sim_i/\tau)}{\sum_{i=0}^{N-1}\exp(sim_i/\tau)}, \label{eq12}
\end{equation}
where $N$ is the number of total classes and $\tau$ represents the temperature parameter of CLIP.

\subsection{Label Augmentation via PN Descriptors}
\label{labelencoder}
Although in-the-wild DFER usually comprises limited categories (\textit{e.g.}, 7 in DFEW~\cite{jiang2020dfew} and FERV39k \cite{wang2022ferv39k}, or 11 in MAFW~\cite{liu2022mafw}), the recognition difficulty does not reduce due to the high inter-class ambiguity (as shown in Tab.~\ref{tab:specific}). Therefore, as stated in Sec.~\ref{pipeline}, class labels are transformed into textual supervision for utilizing their semantic meanings.

While existing CLIP-based DFER models~\cite{zhao2023prompting, tao2024a3ligndferpioneeringcomprehensivedynamic, li2023cliper} mostly focus on enriching the textual descriptions for ground truth labels from a positive view,
in this work, we devise a different label augmentation strategy by extending the original class labels from both positive and negative perspectives.
Specifically, the Positive-Negative (PN) descriptors are derived as follows: \textit{i.e.}, P(ositive): \textit{"A person with an expression of \{Cls\}."}, and N(egative): \textit{"A person with an expression of \textbf{no} \{Cls\}."}.
Correspondingly, the augmented textual supervision $\mathcal{C}$ could contain two different collections, namely $\mathcal{C}_P$ for positive collections and $\mathcal{C}_N$ for negative collections.
Then, both text collections are tokenized and projected into word embeddings obtaining $\mathbf{X}_{T_P}, \mathbf{X}_{T_N}\in\mathbb{R}^{l\times d_{T}}$, where $l$ represents the text length. The inputs are further constructed as:
\begin{equation}
    \mathbf{z}_{T_P}^{(0)}=\mathbf{X}_{T_P}+\mathbf{E}_{T_P}, \quad\mathbf{z}_{T_N}^{(0)}=\mathbf{X}_{T_N}+\mathbf{E}_{T_N}, \label{eq3}
\end{equation}
where $\mathbf{E}$ denotes the positional encoding.

To further encode $ \mathbf{z}_{T_P}^{(0)}$ and $\mathbf{z}_{T_N}^{(0)}$, we resort to the pre-trained textual part of VLM~\cite{radford2021learning}, a model with $L_{T}$ pre-trained transformer layers, devoted by $\{\mathcal{E}_{T}^{(i)}\}_{i=1}^{L_{T}}$.
Keeping the original weights of these well-trained layers,
we introduce trainable lightweight adapters after each frozen layer $\mathcal{E}_{T}^{(j)}$.
denoted as $\{\mathcal{A}_{T_P}^{(j)}\}$ and $\{\mathcal{A}_{T_N}^{(j)}\}$ for positive and negative textual supervision, respectively. 
Then the encoded positive and negative textual features could be obtained via:
\begin{equation}
\begin{aligned}
    \mathbf{z}_{T_P}^{(j)}=\mathcal{E}_{T_P}^{(j)}(\mathcal{A}_{T_P}^{(j)}(\mathbf{z}_{T_P}&^{(j-1)})),\quad\mathbf{z}_{T_N}^{(j)}=\mathcal{E}_{T_N}^{(j)}(\mathcal{A}_{T_N}^{(j)}(\mathbf{z}_{T_N}^{(j-1)})). \label{eq4}
\end{aligned}
\end{equation}
We adopt the basic Adapter structure proposed in~\cite{houlsby2019parameter} for all adapters in our FineCLIPER framework. The structure of the adapter is illustrated in the middle of Fig.~\ref{fig:framework}. Then the final positive and negative text representations can be obtained by:
\begin{equation}
    \mathbf{c}_P=\mathbf{h}_{T}(\mathbf{z}_{T_P,l}^{(L_{T})}),\quad \mathbf{c}_N=\mathbf{h}_{T}(\mathbf{z}_{T_N,l}^{(L_{T})}),
\end{equation}
where $\mathbf{z}_{T, l}^{(L_{T})}$ is the last token of $\mathbf{z}_{T}^{(L_{T})}$ and $\mathbf{h}_{T}$ is a projection layer.

\begin{figure}
    \centering
    \vspace{-0.6em}
   \includegraphics[width=1\linewidth]{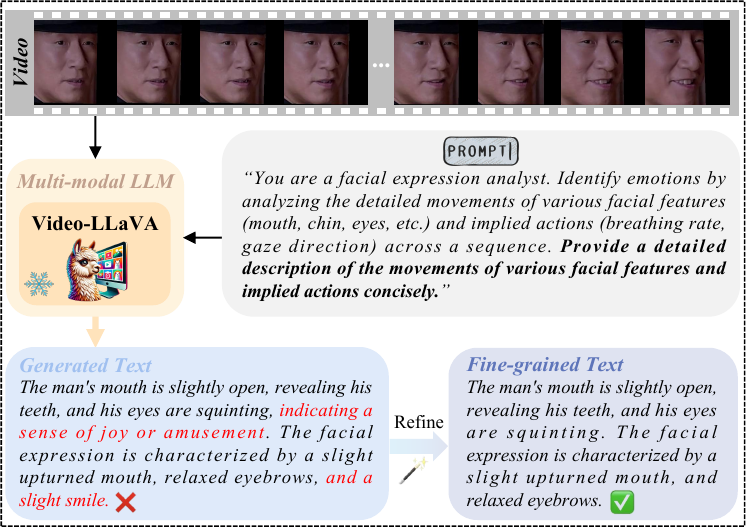}
   \vspace{-2em}
    \captionof{figure}{\label{fig:textgene}
    Fine-grained Text Generation and Refinement.}
    \vspace{-0.8em}
\end{figure}

\subsection{Hierarchical Information Mining}
\label{mining}
Our FineCLIPER adopts a hierarchical manner to mine useful information from:
1) \textit{low semantic level}, where video frames are directly embedded;
2) \textit{middle semantic level}, where face segmentation and landmarks are exploited, and
3) \textit{high semantic level}, where fine-grained descriptions are obtained via MLLM to depict facial dynamics across frames.
Details can be found as follows:

\noindent \textbf{Video Frames Embedding} could provide semantically low-level features since the model operates at pixel-level. To effectively explore the spatial-temporal visual information, we resort to the strong spatial modeling abilities displayed by CLIP and utilize a temporal-expanded version inspired by~\cite{yang2023aim}. 

Formally, given a video clip $V\in\mathbb{R}^{T\times H\times W\times3}$, where $H\times W$ is the spatial size and $T$ is the temporal length.
For $t$-th frame, we spatially divide it into non-overlapping patches $\{\mathbf{P}_{t,i}\}_{i=1}^{M}\in\mathbb{R}^{P^{2}\times3}$, where $M = HW/P^2$. 
These patches are then projected into patch embeddings $\mathbf{X}_{v,t}\in\mathbb{R}^{M \times d}$, where $d$ represents the embedding dimension.
Therefore, the representation for the given video $V$ could be $\mathbf{z}\in\mathbb{R}^{T\times{ M\times d}}$.
After the temporal information undergoes processing by the temporal adapter, the spatially adapted feature can be derived through the following procedure:
\begin{equation}
    \mathbf{z}_{TemV}^{(j)}=\mathcal{E}_{V}^{(j)}(\mathcal{A}_{V}^{(j)}(\mathbf{z}^{(j)})),
\end{equation}

\begin{equation}
    \mathbf{z}_{SpaV}^{(j)}=\mathcal{E}_{V}^{(j)}(\mathcal{A}_{V}^{(j)}(\mathbf{z}_{TemV}^{(j)})), \label{eq5}
\end{equation}
where $\mathbf{z}_{TemV}^{(j)}$ and $\mathbf{z}_{SpaV}^{(j)}$ denotes the temporally and spatially adapted features, respectively.

As a result, the adapter, operating in parallel with the MLP layer, aims to collectively refine the representation of spatiotemporal information. The final feature, scaled by a factor $s$ (set to 0.5 in our framework), can be expressed as follows:
\begin{equation}
    \mathbf{z}_{V}^{(j)}=\mathbf{z}_{SpaV}^{(j)} + {MLP}(LN(\mathbf{z}_{SpaV}^{(j)})) + s\cdot\mathcal{A}_{V}^{(j)}(LN(\mathbf{z}_{SpaV}^{(j)})). \label{eq6}
\end{equation}
Thus, the ultimate video representation at a low semantic level is derived as $\mathbf{v}=\mathbf{h}_{V}(\mathbf{z}_{V}^{(L_{V})})$.

\noindent \textbf{Face Parsing and Landmarks Detection.}
Based on a given frame, we could further mine middle-level semantic information from it.
In our task, as the main part of a frame is mostly human faces, we choose to utilize a powerful
facial analysis model, FaceXFormer~\cite{narayan2024facexformer}, to obtain generalized and robust face representations.
Specifically, we extract the facial segmentation map and perform landmark detection.
Intuitively, the former implies the semantically grouped facial regions, while the latter could provide accurate locations indicating different face parts (\textbf{e.g.}, eyes, nose, etc.)

Specifically, given a specific video clip $V$, the extracted parsing results and landmark maps are represented as $P$ and $L$, respectively.
Following patch embedding, 
both $P$ and $L$ are fed into the corresponding segmentation encoder $\mathbf{E}_P$ and landmark encoder $\mathbf{E}_L$, similar to the operation done for the frame data.
The encoders $\mathbf{E}_P$ and $\mathbf{E}_L$ share weights to collaboratively capture middle-level face semantics. 
Finally, the parsing and landmark representations can be obtained as $\mathbf{p}=\mathbf{h}_{P}(\mathbf{z}_{P}^{(L_{P})})$ and $\mathbf{l}=\mathbf{h}_{L}(\mathbf{z}_{L}^{(L_{L})})$, and $\mathbf{h}_{P}$ and $\mathbf{h}_{L}$ are projection layers for $P$ and $L$, respectively.

\noindent \textbf{Additional Fine-grained Descriptions.} 
In this part, we try to achieve fine-grained details describing the facial dynamics across video frames to serve as high-level semantics.
Specifically,
for each video clip $V$, we adopt Video-LLaVA~\cite{lin2023video}, a MLLM, to generate detailed descriptions under the guidance of an elaborately designed prompt, where the model is asked to play a role as a facial expression analyst to provide details of facial changes, as illustrated in Fig.~\ref{fig:textgene}. 
To elaborate, the provided text prompt raises requirements for the granularity of the descriptions, 
explicitly specifying movements involving various local facial regions.
However, the generated description may include emotion-related words associated with the label or contain some redundant information. 
Hence, we thoroughly refined all generated descriptions to achieve a concise and high-quality summary. 
The refinement works as follows. Initially, we employed a rule-based approach, utilizing pre-configured regular filters to eliminate redundant and irrelevant textual information. Popular text processing tools from the NLTK package were then utilized to remove noise. Subsequently, each data entry will go through manual inspection to filter out abnormal descriptions.

The average number of tokens in our refined descriptions is approximately 35 tokens. However, research \cite{zhang2024long} demonstrates the actual effective length of CLIP's text encoder is even less than 20 tokens. Hence, to better explore the fine-grained description of facial changes, we adopt the text encoder of Long-CLIP~\cite{zhang2024long} as our fine-grained text encoder $\mathbf{E}_F$, which can support text inputs of up to 248 tokens. %
The refined fine-grained description, denoted as $F$, is further tokenized and projected into embeddings $\mathbf{X}_{F}$. Following a procedure similar to the text encoder described in Sec.~\ref{labelencoder}, the input is further constructed as $\mathbf{z}_{F}^{(0)}=\mathbf{X}_{F}+\mathbf{E}_{F}$, where $\mathbf{E}_{F}$ is the positional encoding of $F$.
Subsequently, by feeding it into the projector $\mathbf{h}_{F}$, we could contain the final feature vector of $F$ as: 
$\mathbf{f}=\mathbf{h}_{F}(\mathbf{z}_{F,l}^{(L_{F})})$.

\subsection{Weighted Integration.} 
\label{fusion}
Through the aforementioned semantically hierarchical information mining process, we obtain:
1) low-level video frame feature $\mathbf{v}$,
2) middle-level face parsing features $\mathbf{p}$ and face landmark features $\mathbf{l}$, and
3) high-level fine-grained description features $\mathbf{f}$.
The integration of these features is done using an adaptive fusion strategy. 

Specifically,
given a specific video $V$, the supervision for the $i^{th}$ class is represented by both the positive $\mathbf{c}^i_P$ and negative $\mathbf{c}^i_N$.
Suppose any representation $\mathbf{m}\in\{\mathbf{v}, \mathbf{p}, \mathbf{l}, \mathbf{f}\}$, 
the similarity between $\mathbf{m}$ and $\mathbf{c}_P$, as well as $\mathbf{m}$ and $\mathbf{c}_N$ is defined by calculating the cosine similarity:
\begin{equation}
    sim_{i, \mathbf{m}}^{pos}=\frac{\mathbf{c}_P^i\cdot \mathbf{m}}{\left\|\mathbf{c}_P^i\right\|\left\|\mathbf{m}\right\|},\quad sim_{i, \mathbf{m}}^{neg}=\frac{\mathbf{c}_N^i\cdot \mathbf{m}}{\left\|\mathbf{c}_N^i\right\|\left\|\mathbf{m}\right\|}, \label{eq7}
\end{equation}
is obtained by: $sim_{i,\mathbf{m}} = sim_{i,\mathbf{m}}^{pos} - sim_{i,\mathbf{m}}^{neg}$, which further distinguishes similarity among similar categories 

\begin{figure}
    \centering
   \includegraphics[width=1\linewidth]{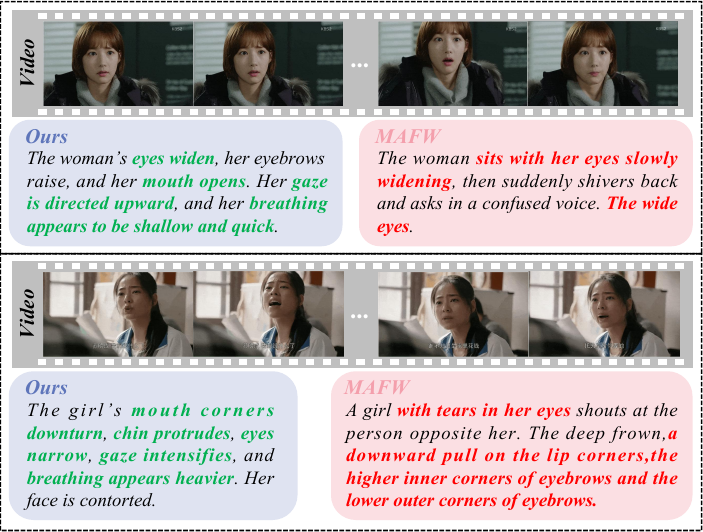}
   \vspace{-2em}
    \captionof{figure}{\label{fig:mafw}
    Comparison of video caption examples between our generated captions and those of the MAFW dataset. Our captions precisely describe facial activities (highlighted in green), in contrast to the MAFW descriptions, which are overly broad and tedious (highlighted in red).} 
    \vspace{-0.8em}
\end{figure}

Then, by finding the max similarity across all the categories, we obtain
$sim_{\mathbf{v}} = max_{i=0}^{N}(sim_{i,\mathbf{v}})$.
Similarly, we could get $sim_{\mathbf{f}}$, $sim_{\mathbf{p}}$, and $sim_{\mathbf{l}}$ following corresponding max-similarity category.
Normalizing these similarities, we obtain the weights corresponding to that representation as:
\begin{equation}
    w_{\mathbf{m}}=\frac{e^{sim_\mathbf{m}}}{e^{sim_\mathbf{v}}+e^{sim_\mathbf{f}}+e^{sim_\mathbf{p}}+e^{sim_\mathbf{l}}}. \label{eq8}
\end{equation}
Such weights could be calculated for $\mathbf{p}, \mathbf{l}, \mathbf{f}$ similarly, resulting in
the corresponding weights $w_{\mathbf{v}}$, $w_{\mathbf{f}}$, $w_{\mathbf{p}}$, and $w_{\mathbf{l}}$. Then the overall multi-modal representation $\mathbf{v}^{mm}$ of Multi-Modal Encoders can be obtained as follows:
\begin{equation}
    \mathbf{v}^{mm}=w_{\mathbf{v}}\cdot\mathbf{v} + w_{\mathbf{f}}\cdot\mathbf{f} + w_{\mathbf{p}}\cdot\mathbf{p} + w_{\mathbf{l}}\cdot\mathbf{l}. \label{eq9}
\end{equation}
where the weights also correspond to the weights of the cross-entropy loss for each modality. Then the overall loss function can thus be expressed as:
\begin{equation}
    \begin{aligned}
    &\mathcal{L}=\frac1{\mathcal{B}}\sum_{i=1}^{\mathcal{B}}(\mathcal{H}(y_i,p(cls_i|\mathbf{v}^{mm}))+
    \\
    &w_{\mathbf{v}}\cdot\mathcal{H}(y_i,p(cls_i|\mathbf{v}))
    +w_{\mathbf{p}}\cdot\mathcal{H}(y_i,p(cls_i|\mathbf{p}))
    \\
    &+w_{\mathbf{l}}\cdot\mathcal{H}(y_i,p(cls_i|\mathbf{l}))+w_{\mathbf{f}}\cdot\mathcal{H}(y_i,p(cls_i|\mathbf{f}))),
    \end{aligned}
\end{equation}
where $\mathcal{B}$ and $\mathcal{H}$ denote the batch size and the cross-entropy loss, respectively.

\section{Experiment}
\begingroup
\setlength{\tabcolsep}{6pt} 
\begin{table*}
  \renewcommand{\arraystretch}{1}
  \centering
  \caption{Comparisons of our FineCLIPER with the state-of-the-art Supervised DFER methods on DFEW, FERV39k, and MAFW. 
  $^{\ast}$: FineCLIPER with face parsing and landmarks modalities; $^{\dagger}$: FineCLIPER with fine-grained text modality.
  The best results are highlighted in \textbf{Bold}, and the second-best \underline{Underlined}.}
  \centering
  \vspace{-1em}
  \small
   \begin{tabular}{lcccccccc}
     \hlineB{2.5}
     \multirow{2}{*}{Method} & \multirow{2}{*}{Backbone} & Tunable & \multicolumn{2}{c}{DFEW}& \multicolumn{2}{c}{FERV39k} & \multicolumn{2}{c}{MAFW}\\
     \cline{4-9}
     & & Param (M) & UAR & WAR & UAR & WAR & UAR & WAR \\
     \hlineB{2}
     EC-STFL (MM'20)~\cite{jiang2020dfew} & C3D / P3D & 78 & 45.35 & 56.51 & - & - & - & -  \\
     Former-DFER (MM'21)~\cite{10.1145/3474085.3475292} & Transformer & 18 & 53.69 & 65.70 & 37.20 & 46.85 & 31.16 & 43.27  \\
     CEFLNet (IS'22)~\cite{liu2022clip} & ResNet-18 & 13 & 51.14 & 65.35 & - & - & - & -  \\
     NR-DFERNet (ArXiv'22)~\cite{li2022nr} & CNN-Transformer & - & 54.21 & 68.19 & 33.99 & 45.97 & - & -  \\
     STT (ArXiv'22)~\cite{ma2022spatio} & ResNet-18 & - & 54.58 & 66.65 & 37.76 & 48.11 & - & -  \\
     DPCNet (MM'22)~\cite{10.1145/3503161.3547865} & ResNet-50 (first 5 layers) & - & 57.11 & 66.32 & - & - & - & -  \\
     T-ESFL (MM'22)~\cite{liu2022mafw} & ResNet-Transformer & - & - & - & - & - & 33.28 & 48.18  \\
     EST (PR'23)~\cite{liu2023expression} & ResNet-18 & 43 & 53.94 & 65.85 & - & - & - & -  \\
     Freq-HD (MM'23)~\cite{10.1145/3581783.3611972} & VGG13-LSTM & - & 46.85 & 55.68 & 33.07 & 45.26 & - & -  \\
     LOGO-Former (ICASSP'23)~\cite{ma2023logo} & ResNet-18 & - & 54.21 & 66.98 & 38.22 & 48.13 & - & -  \\
     IAL (AAAI'23)~\cite{li2023intensity} & ResNet-18 & 19 & 55.71 & 69.24 & 35.82 & 48.54 & - & -  \\
     AEN (CVPRW'23)~\cite{lee2023frame} & ResNet-18 & - & 56.66 & 69.37 & 38.18 & 47.88 & - & -  \\
     M3DFEL (CVPR'23)~\cite{wang2023rethinking} & ResNet-18-3D & - & 56.10 & 69.25 & 35.94 & 47.67 & - & -  \\
     MAE-DFER (MM'23)~\cite{sun2023mae} & ViT-B/16 & 85 & 63.41 & 74.43 & 43.12 & 52.07 & 41.62 & 54.31  \\
     S2D (ArXiv'23)~\cite{chen2023static} & ViT-B/16 & 9 & 65.45 & 74.81 & 43.97 & 46.21 & 43.40 & 52.55  \\
     CLIPER (ArXiv'23)~\cite{li2023cliper} & CLIP-ViT-B/16 & 88 & 57.56 & 70.84 & 41.23 & 51.34 & - & -  \\
     DFER-CLIP (BMVC'23)~\cite{zhao2023prompting} & CLIP-ViT-B/32 & 90 & 59.61 & 71.25 & 41.27 & 51.65 & 39.89 & 52.55  \\
     EmoCLIP (FG'24)~\cite{foteinopoulou_emoclip_2024} & CLIP-ViT-B/32 & - & 58.04 & 62.12 & 31.41 & 36.18 & 34.24 & 41.46  \\
     A$^3$lign-DFER (ArXiv'24)~\cite{tao2024a3ligndferpioneeringcomprehensivedynamic} & CLIP-ViT-L/14 & - & 64.09 & 74.20 & 41.87 & 51.77 & 42.07 & 53.24  \\
     \hline
     \rowcolor{cyan!10}
     FineCLIPER (Ours) & CLIP-ViT-B/16 & 13 & 62.81 & 72.86 & 42.88 & 52.01 & 42.19 & 53.12 \\
     \rowcolor{cyan!10}
     FineCLIPER$^{\ast}$ (Ours) & CLIP-ViT-B/16 & 19 & 64.89 &  \underline{75.05} & \underline{44.15} & 52.12 & 43.02 & \underline{54.69} \\
     \rowcolor{cyan!10}
     FineCLIPER$^{\dagger}$ (Ours) & CLIP-ViT-B/16 & 14 & \underline{65.72} & 75.01 & 43.86 & \underline{53.02} & \underline{43.91} & 54.11 \\
     \rowcolor{cyan!10}
     FineCLIPER$^{\ast}$$^{\dagger}$ (Ours) & CLIP-ViT-B/16 & 20 & \textbf{65.98} & \textbf{76.21} & \textbf{45.22} & \textbf{53.98} & \textbf{45.01} & \textbf{56.91} \\
     \hlineB{2.5}
   \end{tabular}
  \label{tab:1}
\end{table*} 
\endgroup

\begingroup
\setlength{\tabcolsep}{6pt} 
\begin{table*}
\renewcommand{\arraystretch}{1}
  \centering
  \caption{Comparative analyses of recall across various emotion categories: FineCLIPER \textit{vs.} other approaches on DFEW.}
  \centering
  \vspace{-1em}
  \small
   \begin{tabular}{lcccccccc|cc}
     \hlineB{2.5}
     \multirow{2}{*}{Method} & Tunable & \multicolumn{7}{c}{Recall of Each Emotion} & \multicolumn{2}{c}{DFEW}\\
     \cline{3-9} \cline{10-11}
     & Param (M) & Hap. & Sad. & Neu. & Ang. & Sur. & Dis. & Fea. & UAR & WAR \\
     \hlineB{2}
     Former-DFER (MM'21)~\cite{10.1145/3474085.3475292} & 18 & 84.05 & 62.57 & 67.52 & 70.03 & 56.43 & 3.45 & 31.78 & 53.69 & 65.70 \\
     CEFLNet (IS'22)~\cite{liu2022clip} & 13 & 84.00 & 68.00 & 67.00 & 70.00 & 52.00 & 0.00 & 17.00 & 51.14 & 65.35  \\
     NR-DFERNet (ArXiv'22)~\cite{li2022nr} & - & 88.47 & 64.84 & 70.03 & 75.09 & 61.60 & 0.00 & 19.43 & 54.21 & 68.19  \\
     STT (ArXiv'22)~\cite{ma2022spatio} & - & 87.36 & 67.90 & 64.97 & 71.24 & 53.10 & 3.49 & 34.04 & 54.58 & 66.65  \\
     EST (PR'23)~\cite{liu2023expression} & 43 & 86.87 & 66.58 & 67.18 & 71.84 & 47.53 & 5.52 & 28.49 & 53.43 & 65.85  \\
     IAL (AAAI'23)~\cite{li2023intensity} & 19 & 87.95 & 67.21 & 70.10 & 76.06 & 62.22 & 0.00 & 36.44 & 55.71 & 69.24  \\
     M3DFEL (CVPR'23)~\cite{wang2023rethinking} & - & 89.59 & 68.38 & 67.88 & 74.24 & 59.69 & 0.00 & 31.64 & 56.10 & 69.25  \\
     S2D (ArXiv'23)~\cite{chen2023static} & 9 & 93.87 & 83.25 & 75.31 & 84.19 & 64.33 & 0.00 & 37.07 & 62.57 & 75.98  \\
     \hline
     \rowcolor{cyan!10}
     FineCLIPER (Ours)  & 13 & 90.79 & 82.05 & 76.02 & 83.44 & 62.24 & 10.35 & 34.81 & 62.81 & 72.86 \\
     \rowcolor{cyan!10}
     FineCLIPER$^{\ast}$ (Ours) & 19 & 93.04 & 83.90 & 76.58 & 83.66 & 64.63 & 13.79 & 38.66 & 64.89 &  \underline{75.05}  \\
     \rowcolor{cyan!10}
     FineCLIPER$^{\dagger}$ (Ours) & 14 & 93.92 & 83.84 & 76.52 & 84.19 & 63.60 & 20.38 & 37.57 & \underline{65.72} & 75.01 \\
     \rowcolor{cyan!10}
     FineCLIPER$^{\ast}$$^{\dagger}$ (Ours) & 20 & 93.95 & 83.92 & 76.24 & 84.23 & 64.35 & 20.47 & 38.72 & \textbf{65.98} & \textbf{76.21}  \\
     \hlineB{2.5}
   \end{tabular}
  \label{tab:specific}
\end{table*} 
\endgroup

\subsection{Setup}

\noindent \textbf{Datasets and Evaluation.}
Following previous works, we adopt both supervised and zero-shot learning paradigms, evaluating our proposed FineCLIPER together with the baselines on the various in-the-wild DFER datasets, including DFEW~\cite{jiang2020dfew}, FERV39k~\cite{wang2022ferv39k}, and MAFW~\cite{liu2022mafw}. 
We utilize UAR (Unweighted Average Recall) and WAR (Weighted Average Recall) as evaluation metrics for our assessments.
Both DFEW and FERV39k have 7 dynamic facial expression categories to recognize, while MAFW has 11 categories.
It is noteworthy that MAFW dataset comes with video captions for each video, making it a choice for pretraining in zero-shot setting.

\noindent \textbf{Implementation Details.}
All the experiments of our FineCLIPER are built on a CLIP model with the backbone of ViT-B/16 using a single NVIDIA RTX 4090 GPU for fairness and consistency. We process the input by resizing and cropping 16 video frames to a uniform size of 224×224 pixels. The SGD optimizer is employed with an initial learning rate of $3 \times 10^{-4}$. FineCLIPER is trained in an end-to-end manner over 30 epochs with the temperature hyper-parameter $\tau=0.01$.
\vspace{-0.6em}

\subsection{Main Results}
\label{main_results}
\noindent \textbf{Supervised Setting.} The quantitative results in the supervised setting on three standard DFER datasets are depicted in Tab.~\ref{tab:1}. It can be observed that our proposed FineCLIPER achieves state-of-the-art performance compared with other DFER approaches. In addition, our method outperforms all CLIP-based DFER methods with the most lightweight architecture and also the least tunable parameters. Furthermore, we investigate three variants of our FineCLIPER, incorporating face parsing and landmark modalities, along with fine-grained text descriptions of facial changes, which justify the combination of these strategies. 
The superiority of our FineCLIPER is also supported by the substantial improvement in the most challenging category for previous methods, \textit{i.e.}, ``Disgust (Dis.)'', as shown in Tab.~\ref{tab:specific}. It is worth noting that even without the hierarchical information modeling, FineCLIPER, which only has PN descriptors with adapters, still achieves competitive performance. 
This demonstrates the effectiveness of the label augmentation strategy via PN descriptors and the usage of PEFT techniques.
Further ablation studies can be found in Sec.~\ref{ablation}.

\begingroup
\setlength{\tabcolsep}{6pt} 
\begin{table*}
\renewcommand{\arraystretch}{1}
  \centering
  \caption{Comparison with state-of-the-art Zero-Shot DFER methods. $^{\dagger}$: FineCLIPER with fine-grained text modality.}
  \centering
  \vspace{-1em}
  \small
   \begin{tabular}{lcccccccc}
     \hlineB{2.5}
     \multirow{2}{*}{Method} & \multirow{2}{*}{Backbone} & Pre-training & \multicolumn{2}{c}{DFEW}& \multicolumn{2}{c}{FERV39k} & \multicolumn{2}{c}{MAFW}\\
     \cline{4-9}
     & & Dataset & UAR & WAR & UAR & WAR & UAR & WAR \\
     \hlineB{2}
     CLIP (ICML'21)~\cite{radford2021learning} & ViT-B/32 & LAION-400M & 23.34 & 20.07 & 20.99 & 17.09 & 18.42 & 19.16  \\
     FaRL (CVPR'22)~\cite{zheng2022general} & ViT-B/16 & LAION Face-20M & 23.14 & 31.54 & 21.67 & 25.65 & 14.18 & 11.78  \\
     EmoCLIP (FG'24)~\cite{foteinopoulou_emoclip_2024} & CLIP-ViT-B/32 & MAFW (class description) & 22.85 & 24.96 & 39.35 & 41.60 & 24.12 & 24.74  \\
     EmoCLIP (FG'24)~\cite{foteinopoulou_emoclip_2024} & CLIP-ViT-B/32 & MAFW (video caption) & 36.76 & 46.27 & 26.73 & 35.30 & 25.86 & 33.49  \\
     \hline
     \rowcolor{yellow!10}
     FineCLIPER$^{\dagger}$ (Ours) & CLIP-ViT-B/16 & MAFW (video caption) & 47.52 & 57.12 & 34.59 & 42.28 & \underline{34.02} & \underline{40.23} \\
     \rowcolor{cyan!10}
     FineCLIPER$^{\dagger}$ (Ours) & CLIP-ViT-B/16 & MAFW (fine-grained caption) & 52.26 & 62.03 & 39.72 & 46.01 & \textbf{38.77} & \textbf{46.12} \\
     \rowcolor{cyan!10}
     FineCLIPER$^{\dagger}$ (Ours) & CLIP-ViT-B/16 & DFEW (fine-grained caption) & \textbf{57.48} & \textbf{65.45} & \underline{40.10} & \underline{46.91} & - & - \\
     \rowcolor{cyan!10}
     FineCLIPER$^{\dagger}$ (Ours) & CLIP-ViT-B/16 & FERV39k (fine-grained caption) & \underline{55.13} & \underline{63.89} & \textbf{40.79} & \textbf{48.63} & - & - \\
     \hlineB{2.5}
   \end{tabular}
   \vspace{-0.2cm}
  \label{tab:2}
\end{table*} 
\endgroup

\begingroup
\setlength{\tabcolsep}{3.4pt}
\begin{table}
\renewcommand{\arraystretch}{1}
  \centering
  \caption{Ablation on different semantic levels respectively on DFEW, FERV39k, and MAFW.}
  \centering
  \vspace{-1em}
  \small
   \begin{tabular}{lcccccc}
     \hlineB{2}
      & \multicolumn{2}{c}{DFEW}& \multicolumn{2}{c}{FERV39k} & \multicolumn{2}{c}{MAFW}\\
     \cline{2-7}
     & UAR & WAR & UAR & WAR & UAR & WAR \\
     \hlineB{1.5}
     \rowcolor{yellow!10}
     Low-level Only & 62.81 & 72.86 & 42.88 & 52.01 & 42.19 & 53.12 \\
     \rowcolor{yellow!10}
     Middle-level Only& 61.91 & 72.62 & 42.92 & 52.22 & 43.02 & 53.11 \\
     \rowcolor{yellow!10}
     High-level Only & 63.15 & 73.95 & 43.13 & 52.97 & 44.12 & 54.60 \\
     \rowcolor{cyan!10}
     Low-Middle-High (Ours) & \textbf{65.98} & \textbf{76.21} & \textbf{45.22} & \textbf{53.98} & \textbf{45.01} & \textbf{56.91} \\
     \hlineB{2}
   \end{tabular}
   \vspace{-0.2cm}
  \label{tab:level}
\end{table} 
\endgroup

\begingroup
\setlength{\tabcolsep}{5.2pt}
\begin{table}
  \centering
\renewcommand{\arraystretch}{1}
  \caption{Performance of FineCLIPER$^{\ast}$ w.r.t. data from parsing and landmark modalities.}
  \centering
  \vspace{-1em}
  \small
   \begin{tabular}{cc|cccccc}
     \hlineB{2.5}
     \multirow{2}{*}{Parsing} & \multirow{2}{*}{Land.} &  \multicolumn{2}{c}{DFEW}& \multicolumn{2}{c}{FERV39k} & \multicolumn{2}{c}{MAFW}\\
     \cline{3-8}
     & & UAR & WAR & UAR & WAR & UAR & WAR \\
     \hlineB{2}
     \rowcolor{yellow!10}
     \ding{55} & \ding{55} & 62.81 & 72.86 & 42.88 & 52.01 & 42.19 & 53.12  \\
     \rowcolor{yellow!10}
    \ding{51} & \ding{55} & 63.66 & 73.86 & \underline{43.66} & 52.00 & \underline{42.78} & \underline{53.59} \\
    \rowcolor{yellow!10}
     \ding{55} & \ding{51} & \underline{63.71} & \underline{74.16} & 43.53 & \underline{52.08} & 42.56 & 53.16  \\
     \rowcolor{cyan!10}
     \ding{51} & \ding{51} & \textbf{64.89} & \textbf{75.05} & \textbf{44.15} & \textbf{52.12} & \textbf{43.02} & \textbf{54.69}  \\
     \hlineB{2.5}
   \end{tabular}
  \label{tab:3}
\end{table} 
\endgroup

\noindent \textbf{Zero-shot Setting.} 
To assess the generalization ability of FineCLIPER, we perform zero-shot DFER using captions extracted directly from each video (\textit{i.e.}, training on one of the three DFER datasets and then testing on the other two datasets).
Our main baseline is EmoCLIP~\cite{foteinopoulou_emoclip_2024}, which is the first CLIP-based zero-shot DFER model, utilizes the MAFW \cite{liu2022mafw} dataset for pertaining. 
The comparison between captions in MAFW and our generated fine-grained descriptions is shown in Fig.~\ref{fig:mafw}.

Tab.~\ref{tab:2} reports the recognition performance of our FineCLIPER compared with other approaches in the zero-shot DFER setting. Not only did we surpass the previous methods when the pretraining data was consistent, but employing our generated fine-grained captions also led to a significant performance improvement. 
This further demonstrates the effectiveness of the fine-grained description obtained and used by our FineCLIPER, which focuses more on facial changes instead of video scenes (as in MAFW).
In other words, fine-grained descriptions play a pivotal role in guiding the model's attention toward detailed aspects of specific facial regions in the zero-shot setting.

\subsection{Ablation Studies}
\label{ablation}
\noindent \textbf{Performance w.r.t. different level facial features.}
We investigate the effectiveness of using low, middle, or high-level semantics on performance. The results are presented in Tab.~\ref{tab:level}. Notably, although the middle semantic level focuses more on faces, the lack of general visual information poses challenges for vision-language models in data understanding. 
Additionally, we examine the effectiveness of middle-level face semantics obtained through face parsing and landmark detection, with results shown in Tab.~\ref{tab:3}. Comparing rows 1-2 and 1-3, middle-level facial features improve performance. Combining face segmentation and landmarks yields the best results, demonstrating their complementary nature.

\begingroup
\setlength{\tabcolsep}{4.8pt}
\begin{table}
\renewcommand{\arraystretch}{1}
  \centering
  \caption{Performance w.r.t. diverse adapter configurations. $pos$ and $neg$ are positive and negative adapters, respectively.}
  \centering
  \vspace{-1em}
  \small
   \begin{tabular}{cc|cccccc}
     \hlineB{2.5}
     \multirow{2}{*}{Text} & \multirow{2}{*}{Video} &  \multicolumn{2}{c}{DFEW}& \multicolumn{2}{c}{FERV39k} & \multicolumn{2}{c}{MAFW}\\
     \cline{3-8}
     & & UAR & WAR & UAR & WAR & UAR & WAR \\
     \hlineB{2}
     \rowcolor{yellow!10}
     \ding{55} & \ding{55} & 59.61 & 71.25 & 41.27 & 51.65 & 39.89 & 52.55  \\
     \rowcolor{yellow!10}
    \ding{51}$_{pos}$ & \ding{55} & 60.32 & 71.55 & 41.51 & 51.70 & 40.47 & 52.62 \\
    \rowcolor{yellow!10}
    \ding{51}$_{pos+neg}$ & \ding{55} & 61.19 & 71.95 & \underline{42.29} & 51.72 & 40.71 & \underline{52.86} \\
    \rowcolor{yellow!10}
     \ding{55} & \ding{51} & \underline{61.88} & \underline{72.08} & 41.56 & \underline{51.77} & \underline{41.26} & 51.44  \\
     \rowcolor{cyan!10}
     \ding{51}$_{pos+neg}$ & \ding{51} & \textbf{62.81} & \textbf{72.86} & \textbf{42.88} & \textbf{52.01} & \textbf{42.19} & \textbf{53.12}  \\
     \hlineB{2.5}
   \end{tabular}
  \label{tab:4}
\end{table} 
\endgroup

\begingroup
\setlength{\tabcolsep}{6pt}
\begin{table}
\renewcommand{\arraystretch}{1}
  \centering
  \caption{Ablation on Negative Descriptor.}
  \centering
  \vspace{-1em}
  \small
\begin{tabular}{lcccccc}
\hlineB{2}
Negative & \multicolumn{2}{c}{DFEW} & \multicolumn{2}{c}{FERV39k} & \multicolumn{2}{c}{MAFW} \\
\cline{2-7}
Descriptor & UAR & WAR & UAR & WAR & UAR & WAR \\
\hlineB{1.5}
\rowcolor{yellow!10}
"less" & 62.15 & 72.51 & 42.11 & 51.79 & 41.97 & 52.91 \\
\rowcolor{cyan!10}
"no" (Ours) & \textbf{62.81} & \textbf{72.86} & \textbf{42.88} & \textbf{52.01} & \textbf{42.19} & \textbf{53.12} \\
\hlineB{2}
\end{tabular}
  \label{tab:neg}
\end{table} 
\endgroup

\begin{figure}
    \centering
   \includegraphics[width=1\linewidth]{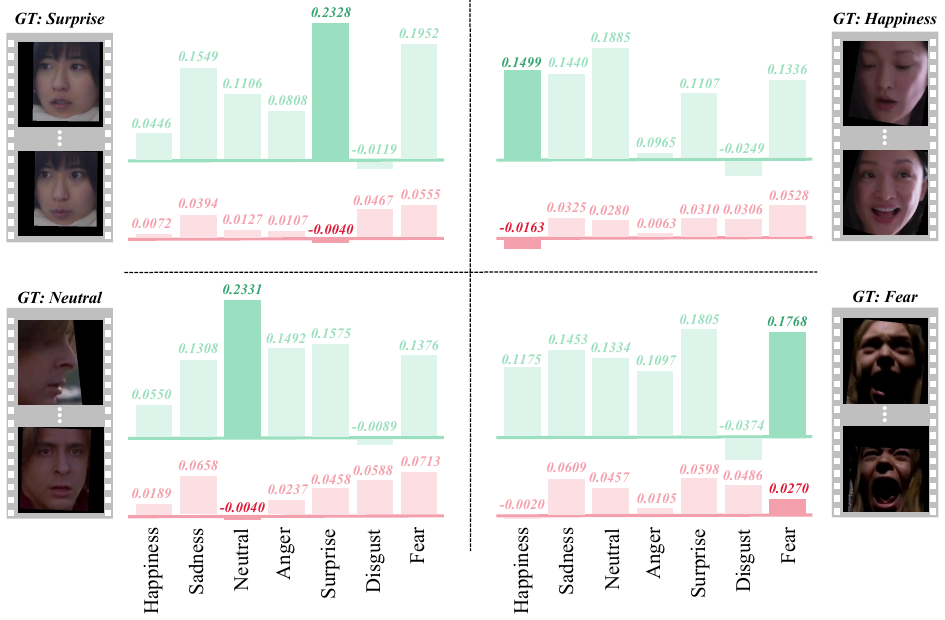}
   \vspace{-1em}
    \captionof{figure}{\label{fig:adapter}
    Visualizations of class-wise cosine similarity values between video and text embeddings in DFEW, where the positive value is in green and the negative one is in red.}
\end{figure}

\begin{figure*}
    \centering
   \includegraphics[width=1\linewidth]{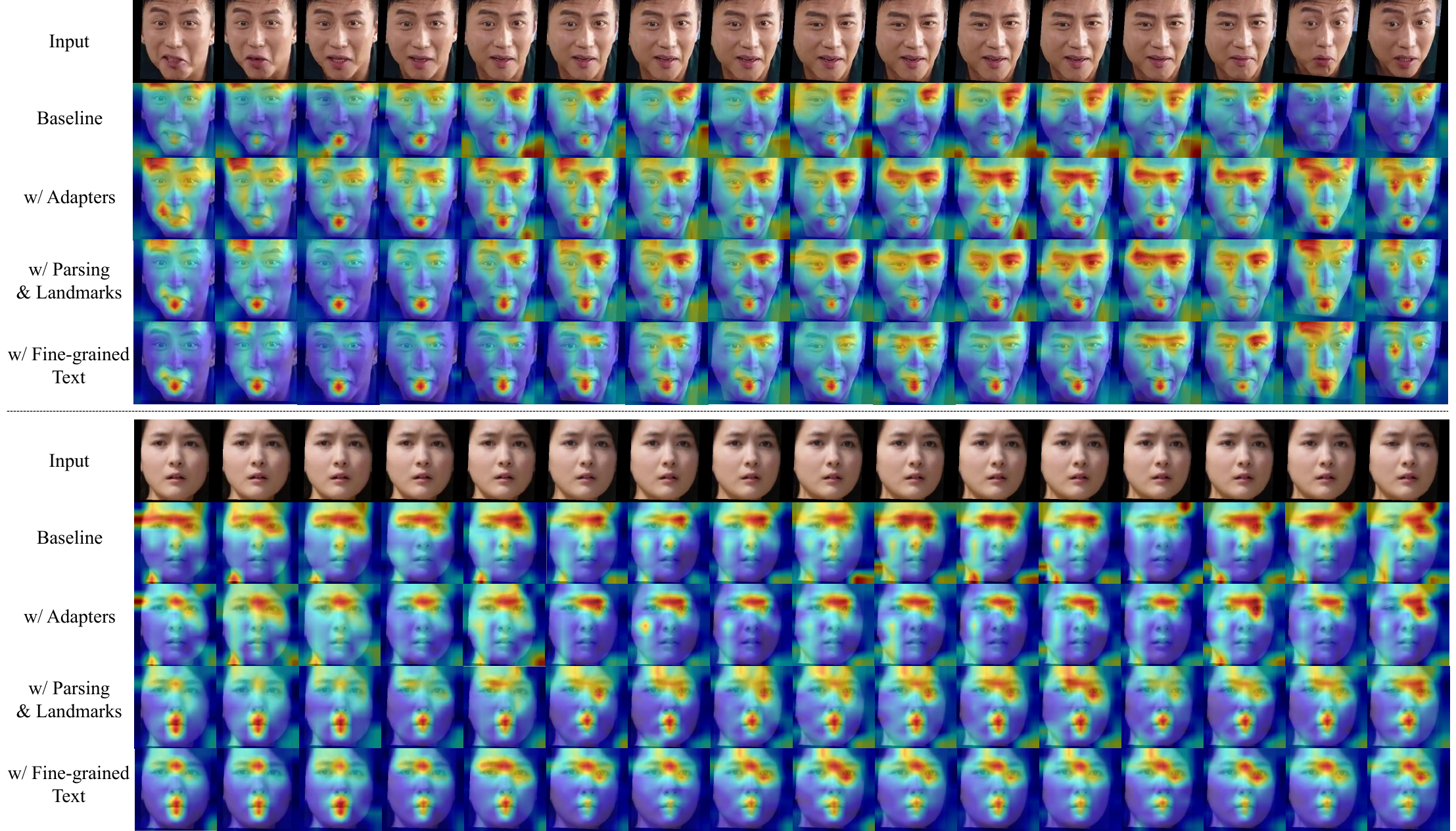}
   \vspace{-2em}
    \captionof{figure}{\label{fig:heatmap}
    Attention visualizations for DFEW w.r.t. two ground-truth expression labels\textemdash `Happiness' (Top) and `Surprise' (Bottom).}
\end{figure*}

\noindent \textbf{Performance w.r.t. label augmentation strategies.}
Since DEFR is a classification task, the supervision typically consists of class labels. However, we extend this supervision to include semantically meaningful textual information, proposing a novel approach that incorporates both positive and negative aspects. The first three rows of Tab.~\ref{tab:4} show the ablations involving this label augmentation strategy. Controlling other variables, our Pos-Neg augmentation achieves the best results across all metrics. To understand the effectiveness of the Pos-Neg descriptors, we visualize the class-wise cosine similarity between video representations and both positive (green) and negative (red) text supervision, as shown in Fig.~\ref{fig:adapter}. This visualization reveals that while positive supervision may sometimes fail (indicated by low positive similarity for some categories), the inclusion of negative supervision helps address these shortcomings.

Considering the complexity of emotions expressed through facial expressions, we further replaced the word \textit{"no"} with \textit{"less"} in the Negative Descriptor, as shown in Tab.~\ref{tab:neg}. The results indicate that using fewer absolute terms leads to poorer performance in the recognition task, which aims to identify prominent emotions.

\noindent \textbf{Performance w.r.t. the usage of trainable adapters.}
We adopted several lightweight trainable adapters in our FineCLIPER to efficiently adapt the ability of large pre-trained models.
The corresponding ablation studies are demonstrated in Tab.~\ref{tab:4}.
We can see that given the same supervision settings (\textit{e.g.}, pose+neg for FineCLIPER),
adding small adaptive modules could effectively boost the performance with only limited trainable parameters (\textit{e.g.}, 20M for all adapters in FineCLIPER).

\noindent \textbf{Effect of each components.}
To validate the effectiveness of each component module in our FineCLIPER, we visualize the attention map of the last transformer block, as shown in Fig.~\ref{fig:heatmap}.
Specifically, we sequentially add components from top to bottom,
including adding the Adapters, using the parsing results and landmarks of faces, as well as using the high-level semantics from the fine-grained descriptions generated by MLLM.
We can see that the model's attention is shrinking to more crucial and concentrated face parts w.r.t. to certain categories.
For example, it focuses on the mouth, eyes, and eyebrows when identifying \textit{Happiness}, 
which aligns well with expression recognition using human vision.
Such visualization results provide a vivid interpretation to explain the superior recognition performance of FineCLIPER.

\section{Conclusion}
Dynamic Facial Expression Recognition (DFER) is vital for understanding human behavior. However, current methods face challenges due to noisy data, neglect of facial dynamics, and confusing categories. To this end,  We propose FineCLIPER, a novel framework with two key innovations:
 1) augmenting class labels with textual PN (Positive-Negative) descriptors to differentiate semantic ambiguity based on the CLIP model's cross-modal latent space; 
 2) employing a hierarchical information mining strategy to mine cues from DFE videos at different semantic levels: \textit{low} (video frame embedding), \textit{middle} (face segmentation masks and landmarks), and \textit{high} (MLLM for detailed descriptions). 
 Additionally, we use Parameter-Efficient Fine-Tuning (PEFT) to adapt all the pre-trained models efficiently.
 FineCLIPER achieves SOTA performance on various datasets with minimal tunable parameters. Detailed ablations and analysis further verify the effectiveness of each design.

\begin{acks}
This work is mainly supported by the National Natural Science Foundation of China (NSFC) under Grant 62306239.
This work is also supported by the Young Talent Fund of Association for Science and Technology in Shaanxi, China.
\end{acks}

\bibliographystyle{ACM-Reference-Format}
\bibliography{sample-base}


\begin{thebibliography}{73}


\ifx \showCODEN    \undefined \def \showCODEN     #1{\unskip}     \fi
\ifx \showDOI      \undefined \def \showDOI       #1{#1}\fi
\ifx \showISBNx    \undefined \def \showISBNx     #1{\unskip}     \fi
\ifx \showISBNxiii \undefined \def \showISBNxiii  #1{\unskip}     \fi
\ifx \showISSN     \undefined \def \showISSN      #1{\unskip}     \fi
\ifx \showLCCN     \undefined \def \showLCCN      #1{\unskip}     \fi
\ifx \shownote     \undefined \def \shownote      #1{#1}          \fi
\ifx \showarticletitle \undefined \def \showarticletitle #1{#1}   \fi
\ifx \showURL      \undefined \def \showURL       {\relax}        \fi
\providecommand\bibfield[2]{#2}
\providecommand\bibinfo[2]{#2}
\providecommand\natexlab[1]{#1}
\providecommand\showeprint[2][]{arXiv:#2}

\bibitem[Baddar and Ro(2019)]%
        {baddar2019mode}
\bibfield{author}{\bibinfo{person}{Wissam~J Baddar} {and} \bibinfo{person}{Yong~Man Ro}.} \bibinfo{year}{2019}\natexlab{}.
\newblock \showarticletitle{Mode variational lstm robust to unseen modes of variation: Application to facial expression recognition}. In \bibinfo{booktitle}{\emph{Proceedings of the AAAI Conference on Artificial Intelligence}}, Vol.~\bibinfo{volume}{33}. \bibinfo{pages}{3215--3223}.
\newblock


\bibitem[Bisogni et~al\mbox{.}(2022)]%
        {9674818}
\bibfield{author}{\bibinfo{person}{Carmen Bisogni}, \bibinfo{person}{Aniello Castiglione}, \bibinfo{person}{Sanoar Hossain}, \bibinfo{person}{Fabio Narducci}, {and} \bibinfo{person}{Saiyed Umer}.} \bibinfo{year}{2022}\natexlab{}.
\newblock \showarticletitle{Impact of Deep Learning Approaches on Facial Expression Recognition in Healthcare Industries}.
\newblock \bibinfo{journal}{\emph{IEEE Transactions on Industrial Informatics}} \bibinfo{volume}{18}, \bibinfo{number}{8} (\bibinfo{year}{2022}), \bibinfo{pages}{5619--5627}.
\newblock
\urldef\tempurl%
\url{https://doi.org/10.1109/TII.2022.3141400}
\showDOI{\tempurl}


\bibitem[Cao et~al\mbox{.}(2014)]%
        {6849440}
\bibfield{author}{\bibinfo{person}{Houwei Cao}, \bibinfo{person}{David~G. Cooper}, \bibinfo{person}{Michael~K. Keutmann}, \bibinfo{person}{Ruben~C. Gur}, \bibinfo{person}{Ani Nenkova}, {and} \bibinfo{person}{Ragini Verma}.} \bibinfo{year}{2014}\natexlab{}.
\newblock \showarticletitle{CREMA-D: Crowd-Sourced Emotional Multimodal Actors Dataset}.
\newblock \bibinfo{journal}{\emph{IEEE Transactions on Affective Computing}} \bibinfo{volume}{5}, \bibinfo{number}{4} (\bibinfo{year}{2014}), \bibinfo{pages}{377--390}.
\newblock
\urldef\tempurl%
\url{https://doi.org/10.1109/TAFFC.2014.2336244}
\showDOI{\tempurl}


\bibitem[Chen et~al\mbox{.}(2024)]%
        {chen2024gaussianvton}
\bibfield{author}{\bibinfo{person}{Haodong Chen}, \bibinfo{person}{Yongle Huang}, \bibinfo{person}{Haojian Huang}, \bibinfo{person}{Xiangsheng Ge}, {and} \bibinfo{person}{Dian Shao}.} \bibinfo{year}{2024}\natexlab{}.
\newblock \showarticletitle{GaussianVTON: 3D Human Virtual Try-ON via Multi-Stage Gaussian Splatting Editing with Image Prompting}.
\newblock \bibinfo{journal}{\emph{arXiv preprint arXiv:2405.07472}} (\bibinfo{year}{2024}).
\newblock


\bibitem[Chen et~al\mbox{.}(2023)]%
        {chen2023static}
\bibfield{author}{\bibinfo{person}{Yin Chen}, \bibinfo{person}{Jia Li}, \bibinfo{person}{Shiguang Shan}, \bibinfo{person}{Meng Wang}, {and} \bibinfo{person}{Richang Hong}.} \bibinfo{year}{2023}\natexlab{}.
\newblock \showarticletitle{From static to dynamic: Adapting landmark-aware image models for facial expression recognition in videos}.
\newblock \bibinfo{journal}{\emph{arXiv preprint arXiv:2312.05447}} (\bibinfo{year}{2023}).
\newblock


\bibitem[Chung et~al\mbox{.}(2014)]%
        {chung2014empirical}
\bibfield{author}{\bibinfo{person}{Junyoung Chung}, \bibinfo{person}{Caglar Gulcehre}, \bibinfo{person}{KyungHyun Cho}, {and} \bibinfo{person}{Yoshua Bengio}.} \bibinfo{year}{2014}\natexlab{}.
\newblock \showarticletitle{Empirical evaluation of gated recurrent neural networks on sequence modeling}.
\newblock \bibinfo{journal}{\emph{arXiv preprint arXiv:1412.3555}} (\bibinfo{year}{2014}).
\newblock


\bibitem[Dong et~al\mbox{.}(2022)]%
        {dong2022dreamartist}
\bibfield{author}{\bibinfo{person}{Ziyi Dong}, \bibinfo{person}{Pengxu Wei}, {and} \bibinfo{person}{Liang Lin}.} \bibinfo{year}{2022}\natexlab{}.
\newblock \showarticletitle{DreamArtist: Towards Controllable One-Shot Text-to-Image Generation via Positive-Negative Prompt-Tuning}.
\newblock \bibinfo{journal}{\emph{arXiv preprint arXiv:2211.11337}} (\bibinfo{year}{2022}).
\newblock


\bibitem[Dosovitskiy et~al\mbox{.}(2020)]%
        {dosovitskiy2020image}
\bibfield{author}{\bibinfo{person}{Alexey Dosovitskiy}, \bibinfo{person}{Lucas Beyer}, \bibinfo{person}{Alexander Kolesnikov}, \bibinfo{person}{Dirk Weissenborn}, \bibinfo{person}{Xiaohua Zhai}, \bibinfo{person}{Thomas Unterthiner}, \bibinfo{person}{Mostafa Dehghani}, \bibinfo{person}{Matthias Minderer}, \bibinfo{person}{Georg Heigold}, \bibinfo{person}{Sylvain Gelly}, {et~al\mbox{.}}} \bibinfo{year}{2020}\natexlab{}.
\newblock \showarticletitle{An image is worth 16x16 words: Transformers for image recognition at scale}.
\newblock \bibinfo{journal}{\emph{arXiv preprint arXiv:2010.11929}} (\bibinfo{year}{2020}).
\newblock


\bibitem[Ebrahimi~Kahou et~al\mbox{.}(2015)]%
        {ebrahimi2015recurrent}
\bibfield{author}{\bibinfo{person}{Samira Ebrahimi~Kahou}, \bibinfo{person}{Vincent Michalski}, \bibinfo{person}{Kishore Konda}, \bibinfo{person}{Roland Memisevic}, {and} \bibinfo{person}{Christopher Pal}.} \bibinfo{year}{2015}\natexlab{}.
\newblock \showarticletitle{Recurrent neural networks for emotion recognition in video}. In \bibinfo{booktitle}{\emph{Proceedings of the 2015 ACM on international conference on multimodal interaction}}. \bibinfo{pages}{467--474}.
\newblock


\bibitem[Fan et~al\mbox{.}(2016)]%
        {fan2016video}
\bibfield{author}{\bibinfo{person}{Yin Fan}, \bibinfo{person}{Xiangju Lu}, \bibinfo{person}{Dian Li}, {and} \bibinfo{person}{Yuanliu Liu}.} \bibinfo{year}{2016}\natexlab{}.
\newblock \showarticletitle{Video-based emotion recognition using CNN-RNN and C3D hybrid networks}. In \bibinfo{booktitle}{\emph{Proceedings of the 18th ACM international conference on multimodal interaction}}. \bibinfo{pages}{445--450}.
\newblock


\bibitem[Foteinopoulou and Patras(2024)]%
        {foteinopoulou_emoclip_2024}
\bibfield{author}{\bibinfo{person}{Niki~Maria Foteinopoulou} {and} \bibinfo{person}{Ioannis Patras}.} \bibinfo{year}{2024}\natexlab{}.
\newblock \showarticletitle{{EmoCLIP}: {A} {Vision}-{Language} {Method} for {Zero}-{Shot} {Video} {Facial} {Expression} {Recognition}}. In \bibinfo{booktitle}{\emph{The 18th IEEE International Conference on Automatic Face and Gesture Recognition}}.
\newblock


\bibitem[Hara et~al\mbox{.}(2018)]%
        {hara2018can}
\bibfield{author}{\bibinfo{person}{Kensho Hara}, \bibinfo{person}{Hirokatsu Kataoka}, {and} \bibinfo{person}{Yutaka Satoh}.} \bibinfo{year}{2018}\natexlab{}.
\newblock \showarticletitle{Can spatiotemporal 3d cnns retrace the history of 2d cnns and imagenet?}. In \bibinfo{booktitle}{\emph{Proceedings of the IEEE conference on Computer Vision and Pattern Recognition}}. \bibinfo{pages}{6546--6555}.
\newblock


\bibitem[Hochreiter and Schmidhuber(1997)]%
        {hochreiter1997long}
\bibfield{author}{\bibinfo{person}{Sepp Hochreiter} {and} \bibinfo{person}{J{\"u}rgen Schmidhuber}.} \bibinfo{year}{1997}\natexlab{}.
\newblock \showarticletitle{Long short-term memory}.
\newblock \bibinfo{journal}{\emph{Neural computation}} \bibinfo{volume}{9}, \bibinfo{number}{8} (\bibinfo{year}{1997}), \bibinfo{pages}{1735--1780}.
\newblock


\bibitem[Houlsby et~al\mbox{.}(2019)]%
        {houlsby2019parameter}
\bibfield{author}{\bibinfo{person}{Neil Houlsby}, \bibinfo{person}{Andrei Giurgiu}, \bibinfo{person}{Stanislaw Jastrzebski}, \bibinfo{person}{Bruna Morrone}, \bibinfo{person}{Quentin De~Laroussilhe}, \bibinfo{person}{Andrea Gesmundo}, \bibinfo{person}{Mona Attariyan}, {and} \bibinfo{person}{Sylvain Gelly}.} \bibinfo{year}{2019}\natexlab{}.
\newblock \showarticletitle{Parameter-efficient transfer learning for NLP}. In \bibinfo{booktitle}{\emph{International conference on machine learning}}. PMLR, \bibinfo{pages}{2790--2799}.
\newblock


\bibitem[Huang et~al\mbox{.}(2024)]%
        {huang2024crest}
\bibfield{author}{\bibinfo{person}{Haojian Huang}, \bibinfo{person}{Xiaozhen Qiao}, \bibinfo{person}{Zhuo Chen}, \bibinfo{person}{Haodong Chen}, \bibinfo{person}{Bingyu Li}, \bibinfo{person}{Zhe Sun}, \bibinfo{person}{Mulin Chen}, {and} \bibinfo{person}{Xuelong Li}.} \bibinfo{year}{2024}\natexlab{}.
\newblock \showarticletitle{CREST: Cross-modal Resonance through Evidential Deep Learning for Enhanced Zero-Shot Learning}.
\newblock \bibinfo{journal}{\emph{arXiv preprint arXiv:2404.09640}} (\bibinfo{year}{2024}).
\newblock


\bibitem[Hundt et~al\mbox{.}(2024)]%
        {hundt2024love}
\bibfield{author}{\bibinfo{person}{Andrew Hundt}, \bibinfo{person}{Gabrielle Ohlson}, \bibinfo{person}{Pieter Wolfert}, \bibinfo{person}{Lux Miranda}, \bibinfo{person}{Sophia Zhu}, {and} \bibinfo{person}{Katie Winkle}.} \bibinfo{year}{2024}\natexlab{}.
\newblock \showarticletitle{Love, joy, and autism robots: A metareview and provocatype}.
\newblock \bibinfo{journal}{\emph{arXiv preprint arXiv:2403.05098}} (\bibinfo{year}{2024}).
\newblock


\bibitem[Jiang et~al\mbox{.}(2020)]%
        {jiang2020dfew}
\bibfield{author}{\bibinfo{person}{Xingxun Jiang}, \bibinfo{person}{Yuan Zong}, \bibinfo{person}{Wenming Zheng}, \bibinfo{person}{Chuangao Tang}, \bibinfo{person}{Wanchuang Xia}, \bibinfo{person}{Cheng Lu}, {and} \bibinfo{person}{Jiateng Liu}.} \bibinfo{year}{2020}\natexlab{}.
\newblock \showarticletitle{Dfew: A large-scale database for recognizing dynamic facial expressions in the wild}. In \bibinfo{booktitle}{\emph{Proceedings of the 28th ACM international conference on multimedia}}. \bibinfo{pages}{2881--2889}.
\newblock


\bibitem[Jiang et~al\mbox{.}(2024)]%
        {jiang2024effectiveness}
\bibfield{author}{\bibinfo{person}{Yao Jiang}, \bibinfo{person}{Xinyu Yan}, \bibinfo{person}{Ge-Peng Ji}, \bibinfo{person}{Keren Fu}, \bibinfo{person}{Meijun Sun}, \bibinfo{person}{Huan Xiong}, \bibinfo{person}{Deng-Ping Fan}, {and} \bibinfo{person}{Fahad~Shahbaz Khan}.} \bibinfo{year}{2024}\natexlab{}.
\newblock \showarticletitle{Effectiveness assessment of recent large vision-language models}.
\newblock \bibinfo{journal}{\emph{arXiv preprint arXiv:2403.04306}} (\bibinfo{year}{2024}).
\newblock


\bibitem[Kollias and Zafeiriou(2020)]%
        {kollias2020exploiting}
\bibfield{author}{\bibinfo{person}{Dimitrios Kollias} {and} \bibinfo{person}{Stefanos Zafeiriou}.} \bibinfo{year}{2020}\natexlab{}.
\newblock \showarticletitle{Exploiting multi-cnn features in cnn-rnn based dimensional emotion recognition on the omg in-the-wild dataset}.
\newblock \bibinfo{journal}{\emph{IEEE Transactions on Affective Computing}} \bibinfo{volume}{12}, \bibinfo{number}{3} (\bibinfo{year}{2020}), \bibinfo{pages}{595--606}.
\newblock


\bibitem[Kossaifi et~al\mbox{.}(2020)]%
        {kossaifi2020factorized}
\bibfield{author}{\bibinfo{person}{Jean Kossaifi}, \bibinfo{person}{Antoine Toisoul}, \bibinfo{person}{Adrian Bulat}, \bibinfo{person}{Yannis Panagakis}, \bibinfo{person}{Timothy~M Hospedales}, {and} \bibinfo{person}{Maja Pantic}.} \bibinfo{year}{2020}\natexlab{}.
\newblock \showarticletitle{Factorized higher-order cnns with an application to spatio-temporal emotion estimation}. In \bibinfo{booktitle}{\emph{Proceedings of the IEEE/CVF conference on computer vision and pattern recognition}}. \bibinfo{pages}{6060--6069}.
\newblock


\bibitem[Kumar et~al\mbox{.}(2024)]%
        {kumar2024measuring}
\bibfield{author}{\bibinfo{person}{Puneet Kumar}, \bibinfo{person}{Alexander Vedernikov}, {and} \bibinfo{person}{Xiaobai Li}.} \bibinfo{year}{2024}\natexlab{}.
\newblock \showarticletitle{Measuring Non-Typical Emotions for Mental Health: A Survey of Computational Approaches}.
\newblock \bibinfo{journal}{\emph{arXiv preprint arXiv:2403.08824}} (\bibinfo{year}{2024}).
\newblock


\bibitem[Ladak et~al\mbox{.}(2024)]%
        {ladak2024artificial}
\bibfield{author}{\bibinfo{person}{Ali Ladak}, \bibinfo{person}{Jamie Harris}, {and} \bibinfo{person}{Jacy~Reese Anthis}.} \bibinfo{year}{2024}\natexlab{}.
\newblock \showarticletitle{Which Artificial Intelligences Do People Care About Most? A Conjoint Experiment on Moral Consideration}. In \bibinfo{booktitle}{\emph{Proceedings of the CHI Conference on Human Factors in Computing Systems}}. \bibinfo{pages}{1--11}.
\newblock


\bibitem[Lee et~al\mbox{.}(2023)]%
        {lee2023frame}
\bibfield{author}{\bibinfo{person}{Bokyeung Lee}, \bibinfo{person}{Hyunuk Shin}, \bibinfo{person}{Bonhwa Ku}, {and} \bibinfo{person}{Hanseok Ko}.} \bibinfo{year}{2023}\natexlab{}.
\newblock \showarticletitle{Frame level emotion guided dynamic facial expression recognition with emotion grouping}. In \bibinfo{booktitle}{\emph{Proceedings of the IEEE/CVF Conference on Computer Vision and Pattern Recognition}}. \bibinfo{pages}{5680--5690}.
\newblock


\bibitem[Lee et~al\mbox{.}(2019)]%
        {lee2019context}
\bibfield{author}{\bibinfo{person}{Jiyoung Lee}, \bibinfo{person}{Seungryong Kim}, \bibinfo{person}{Sunok Kim}, \bibinfo{person}{Jungin Park}, {and} \bibinfo{person}{Kwanghoon Sohn}.} \bibinfo{year}{2019}\natexlab{}.
\newblock \showarticletitle{Context-aware emotion recognition networks}. In \bibinfo{booktitle}{\emph{Proceedings of the IEEE/CVF international conference on computer vision}}. \bibinfo{pages}{10143--10152}.
\newblock


\bibitem[Li et~al\mbox{.}(2023b)]%
        {li2023cliper}
\bibfield{author}{\bibinfo{person}{Hanting Li}, \bibinfo{person}{Hongjing Niu}, \bibinfo{person}{Zhaoqing Zhu}, {and} \bibinfo{person}{Feng Zhao}.} \bibinfo{year}{2023}\natexlab{b}.
\newblock \showarticletitle{Cliper: A unified vision-language framework for in-the-wild facial expression recognition}.
\newblock \bibinfo{journal}{\emph{arXiv preprint arXiv:2303.00193}} (\bibinfo{year}{2023}).
\newblock


\bibitem[Li et~al\mbox{.}(2023c)]%
        {li2023intensity}
\bibfield{author}{\bibinfo{person}{Hanting Li}, \bibinfo{person}{Hongjing Niu}, \bibinfo{person}{Zhaoqing Zhu}, {and} \bibinfo{person}{Feng Zhao}.} \bibinfo{year}{2023}\natexlab{c}.
\newblock \showarticletitle{Intensity-aware loss for dynamic facial expression recognition in the wild}. In \bibinfo{booktitle}{\emph{Proceedings of the AAAI Conference on Artificial Intelligence}}, Vol.~\bibinfo{volume}{37}. \bibinfo{pages}{67--75}.
\newblock


\bibitem[Li et~al\mbox{.}(2022)]%
        {li2022nr}
\bibfield{author}{\bibinfo{person}{Hanting Li}, \bibinfo{person}{Mingzhe Sui}, \bibinfo{person}{Zhaoqing Zhu}, {et~al\mbox{.}}} \bibinfo{year}{2022}\natexlab{}.
\newblock \showarticletitle{Nr-dfernet: Noise-robust network for dynamic facial expression recognition}.
\newblock \bibinfo{journal}{\emph{arXiv preprint arXiv:2206.04975}} (\bibinfo{year}{2022}).
\newblock


\bibitem[Li et~al\mbox{.}(2023a)]%
        {2023videochat}
\bibfield{author}{\bibinfo{person}{Kunchang Li}, \bibinfo{person}{Yinan He}, \bibinfo{person}{Yi Wang}, \bibinfo{person}{Yizhuo Li}, \bibinfo{person}{Wenhai Wang}, \bibinfo{person}{Ping Luo}, \bibinfo{person}{Yali Wang}, \bibinfo{person}{Limin Wang}, {and} \bibinfo{person}{Yu Qiao}.} \bibinfo{year}{2023}\natexlab{a}.
\newblock \showarticletitle{VideoChat: Chat-Centric Video Understanding}.
\newblock \bibinfo{journal}{\emph{arXiv preprint arXiv:2305.06355}} (\bibinfo{year}{2023}).
\newblock


\bibitem[Li et~al\mbox{.}(2024)]%
        {li2024mvbench}
\bibfield{author}{\bibinfo{person}{Kunchang Li}, \bibinfo{person}{Yali Wang}, \bibinfo{person}{Yinan He}, \bibinfo{person}{Yizhuo Li}, \bibinfo{person}{Yi Wang}, \bibinfo{person}{Yi Liu}, \bibinfo{person}{Zun Wang}, \bibinfo{person}{Jilan Xu}, \bibinfo{person}{Guo Chen}, \bibinfo{person}{Ping Luo}, {et~al\mbox{.}}} \bibinfo{year}{2024}\natexlab{}.
\newblock \showarticletitle{Mvbench: A comprehensive multi-modal video understanding benchmark}. In \bibinfo{booktitle}{\emph{Proceedings of the IEEE/CVF Conference on Computer Vision and Pattern Recognition}}. \bibinfo{pages}{22195--22206}.
\newblock


\bibitem[Li and Deng(2022)]%
        {Li_2022}
\bibfield{author}{\bibinfo{person}{Shan Li} {and} \bibinfo{person}{Weihong Deng}.} \bibinfo{year}{2022}\natexlab{}.
\newblock \showarticletitle{Deep Facial Expression Recognition: A Survey}.
\newblock \bibinfo{journal}{\emph{IEEE Transactions on Affective Computing}} \bibinfo{volume}{13}, \bibinfo{number}{3} (\bibinfo{date}{July} \bibinfo{year}{2022}), \bibinfo{pages}{1195–1215}.
\newblock
\showISSN{2371-9850}
\urldef\tempurl%
\url{https://doi.org/10.1109/taffc.2020.2981446}
\showDOI{\tempurl}


\bibitem[Lin et~al\mbox{.}(2023)]%
        {lin2023video}
\bibfield{author}{\bibinfo{person}{Bin Lin}, \bibinfo{person}{Bin Zhu}, \bibinfo{person}{Yang Ye}, \bibinfo{person}{Munan Ning}, \bibinfo{person}{Peng Jin}, {and} \bibinfo{person}{Li Yuan}.} \bibinfo{year}{2023}\natexlab{}.
\newblock \showarticletitle{Video-llava: Learning united visual representation by alignment before projection}.
\newblock \bibinfo{journal}{\emph{arXiv preprint arXiv:2311.10122}} (\bibinfo{year}{2023}).
\newblock


\bibitem[Liu et~al\mbox{.}(2022a)]%
        {liu2022mafw}
\bibfield{author}{\bibinfo{person}{Yuanyuan Liu}, \bibinfo{person}{Wei Dai}, \bibinfo{person}{Chuanxu Feng}, \bibinfo{person}{Wenbin Wang}, \bibinfo{person}{Guanghao Yin}, \bibinfo{person}{Jiabei Zeng}, {and} \bibinfo{person}{Shiguang Shan}.} \bibinfo{year}{2022}\natexlab{a}.
\newblock \showarticletitle{Mafw: A large-scale, multi-modal, compound affective database for dynamic facial expression recognition in the wild}. In \bibinfo{booktitle}{\emph{Proceedings of the 30th ACM International Conference on Multimedia}}. \bibinfo{pages}{24--32}.
\newblock


\bibitem[Liu et~al\mbox{.}(2022b)]%
        {liu2022clip}
\bibfield{author}{\bibinfo{person}{Yuanyuan Liu}, \bibinfo{person}{Chuanxu Feng}, \bibinfo{person}{Xiaohui Yuan}, \bibinfo{person}{Lin Zhou}, \bibinfo{person}{Wenbin Wang}, \bibinfo{person}{Jie Qin}, {and} \bibinfo{person}{Zhongwen Luo}.} \bibinfo{year}{2022}\natexlab{b}.
\newblock \showarticletitle{Clip-aware expressive feature learning for video-based facial expression recognition}.
\newblock \bibinfo{journal}{\emph{Information Sciences}}  \bibinfo{volume}{598} (\bibinfo{year}{2022}), \bibinfo{pages}{182--195}.
\newblock


\bibitem[Liu et~al\mbox{.}(2023)]%
        {liu2023expression}
\bibfield{author}{\bibinfo{person}{Yuanyuan Liu}, \bibinfo{person}{Wenbin Wang}, \bibinfo{person}{Chuanxu Feng}, \bibinfo{person}{Haoyu Zhang}, \bibinfo{person}{Zhe Chen}, {and} \bibinfo{person}{Yibing Zhan}.} \bibinfo{year}{2023}\natexlab{}.
\newblock \showarticletitle{Expression snippet transformer for robust video-based facial expression recognition}.
\newblock \bibinfo{journal}{\emph{Pattern Recognition}}  \bibinfo{volume}{138} (\bibinfo{year}{2023}), \bibinfo{pages}{109368}.
\newblock


\bibitem[Liu et~al\mbox{.}(2017)]%
        {8039024}
\bibfield{author}{\bibinfo{person}{Zhentao Liu}, \bibinfo{person}{Min Wu}, \bibinfo{person}{Weihua Cao}, \bibinfo{person}{Luefeng Chen}, \bibinfo{person}{Jianping Xu}, \bibinfo{person}{Ri Zhang}, \bibinfo{person}{Mengtian Zhou}, {and} \bibinfo{person}{Junwei Mao}.} \bibinfo{year}{2017}\natexlab{}.
\newblock \showarticletitle{A facial expression emotion recognition based human-robot interaction system}.
\newblock \bibinfo{journal}{\emph{IEEE/CAA Journal of Automatica Sinica}} \bibinfo{volume}{4}, \bibinfo{number}{4} (\bibinfo{year}{2017}), \bibinfo{pages}{668--676}.
\newblock
\urldef\tempurl%
\url{https://doi.org/10.1109/JAS.2017.7510622}
\showDOI{\tempurl}


\bibitem[Livingstone and Russo(2018a)]%
        {10.1371/journal.pone.0196391}
\bibfield{author}{\bibinfo{person}{Steven~R. Livingstone} {and} \bibinfo{person}{Frank~A. Russo}.} \bibinfo{year}{2018}\natexlab{a}.
\newblock \showarticletitle{The Ryerson Audio-Visual Database of Emotional Speech and Song (RAVDESS): A dynamic, multimodal set of facial and vocal expressions in North American English}.
\newblock \bibinfo{journal}{\emph{PLOS ONE}} \bibinfo{volume}{13}, \bibinfo{number}{5} (\bibinfo{date}{05} \bibinfo{year}{2018}), \bibinfo{pages}{1--35}.
\newblock
\urldef\tempurl%
\url{https://doi.org/10.1371/journal.pone.0196391}
\showDOI{\tempurl}


\bibitem[Livingstone and Russo(2018b)]%
        {Livingstone2018TheRA}
\bibfield{author}{\bibinfo{person}{Steven~R. Livingstone} {and} \bibinfo{person}{Frank~A. Russo}.} \bibinfo{year}{2018}\natexlab{b}.
\newblock \showarticletitle{The Ryerson Audio-Visual Database of Emotional Speech and Song (RAVDESS): A dynamic, multimodal set of facial and vocal expressions in North American English}.
\newblock \bibinfo{journal}{\emph{PLoS ONE}}  \bibinfo{volume}{13} (\bibinfo{year}{2018}).
\newblock
\urldef\tempurl%
\url{https://api.semanticscholar.org/CorpusID:21704094}
\showURL{%
\tempurl}


\bibitem[Ma et~al\mbox{.}(2022)]%
        {ma2022spatio}
\bibfield{author}{\bibinfo{person}{Fuyan Ma}, \bibinfo{person}{Bin Sun}, {and} \bibinfo{person}{Shutao Li}.} \bibinfo{year}{2022}\natexlab{}.
\newblock \showarticletitle{Spatio-temporal transformer for dynamic facial expression recognition in the wild}.
\newblock \bibinfo{journal}{\emph{arXiv preprint arXiv:2205.04749}} (\bibinfo{year}{2022}).
\newblock


\bibitem[Ma et~al\mbox{.}(2023)]%
        {ma2023logo}
\bibfield{author}{\bibinfo{person}{Fuyan Ma}, \bibinfo{person}{Bin Sun}, {and} \bibinfo{person}{Shutao Li}.} \bibinfo{year}{2023}\natexlab{}.
\newblock \showarticletitle{Logo-former: Local-global spatio-temporal transformer for dynamic facial expression recognition}. In \bibinfo{booktitle}{\emph{ICASSP 2023-2023 IEEE International Conference on Acoustics, Speech and Signal Processing (ICASSP)}}. IEEE, \bibinfo{pages}{1--5}.
\newblock


\bibitem[Martin et~al\mbox{.}(2006)]%
        {1623803}
\bibfield{author}{\bibinfo{person}{O. Martin}, \bibinfo{person}{I. Kotsia}, \bibinfo{person}{B. Macq}, {and} \bibinfo{person}{I. Pitas}.} \bibinfo{year}{2006}\natexlab{}.
\newblock \showarticletitle{The eNTERFACE' 05 Audio-Visual Emotion Database}. In \bibinfo{booktitle}{\emph{22nd International Conference on Data Engineering Workshops (ICDEW'06)}}. \bibinfo{pages}{8--8}.
\newblock
\urldef\tempurl%
\url{https://doi.org/10.1109/ICDEW.2006.145}
\showDOI{\tempurl}


\bibitem[Miyake et~al\mbox{.}(2023)]%
        {miyake2023negative}
\bibfield{author}{\bibinfo{person}{Daiki Miyake}, \bibinfo{person}{Akihiro Iohara}, \bibinfo{person}{Yu Saito}, {and} \bibinfo{person}{Toshiyuki Tanaka}.} \bibinfo{year}{2023}\natexlab{}.
\newblock \showarticletitle{Negative-prompt inversion: Fast image inversion for editing with text-guided diffusion models}.
\newblock \bibinfo{journal}{\emph{arXiv preprint arXiv:2305.16807}} (\bibinfo{year}{2023}).
\newblock


\bibitem[Mollahosseini et~al\mbox{.}(2019)]%
        {Mollahosseini_2019}
\bibfield{author}{\bibinfo{person}{Ali Mollahosseini}, \bibinfo{person}{Behzad Hasani}, {and} \bibinfo{person}{Mohammad~H. Mahoor}.} \bibinfo{year}{2019}\natexlab{}.
\newblock \showarticletitle{AffectNet: A Database for Facial Expression, Valence, and Arousal Computing in the Wild}.
\newblock \bibinfo{journal}{\emph{IEEE Transactions on Affective Computing}} \bibinfo{volume}{10}, \bibinfo{number}{1} (\bibinfo{date}{Jan.} \bibinfo{year}{2019}), \bibinfo{pages}{18–31}.
\newblock
\showISSN{2371-9850}
\urldef\tempurl%
\url{https://doi.org/10.1109/taffc.2017.2740923}
\showDOI{\tempurl}


\bibitem[Narayan et~al\mbox{.}(2024)]%
        {narayan2024facexformer}
\bibfield{author}{\bibinfo{person}{Kartik Narayan}, \bibinfo{person}{Vibashan VS}, \bibinfo{person}{Rama Chellappa}, {and} \bibinfo{person}{Vishal~M Patel}.} \bibinfo{year}{2024}\natexlab{}.
\newblock \showarticletitle{FaceXFormer: A Unified Transformer for Facial Analysis}.
\newblock \bibinfo{journal}{\emph{arXiv preprint arXiv:2403.12960}} (\bibinfo{year}{2024}).
\newblock


\bibitem[Radford et~al\mbox{.}(2021)]%
        {radford2021learning}
\bibfield{author}{\bibinfo{person}{Alec Radford}, \bibinfo{person}{Jong~Wook Kim}, \bibinfo{person}{Chris Hallacy}, \bibinfo{person}{Aditya Ramesh}, \bibinfo{person}{Gabriel Goh}, \bibinfo{person}{Sandhini Agarwal}, \bibinfo{person}{Girish Sastry}, \bibinfo{person}{Amanda Askell}, \bibinfo{person}{Pamela Mishkin}, \bibinfo{person}{Jack Clark}, {et~al\mbox{.}}} \bibinfo{year}{2021}\natexlab{}.
\newblock \showarticletitle{Learning transferable visual models from natural language supervision}. In \bibinfo{booktitle}{\emph{International conference on machine learning}}. PMLR, \bibinfo{pages}{8748--8763}.
\newblock


\bibitem[Sarkar et~al\mbox{.}(2023)]%
        {sarkar2023towards}
\bibfield{author}{\bibinfo{person}{Surjodeep Sarkar}, \bibinfo{person}{Manas Gaur}, \bibinfo{person}{L Chen}, \bibinfo{person}{Muskan Garg}, \bibinfo{person}{Biplav Srivastava}, {and} \bibinfo{person}{Bhaktee Dongaonkar}.} \bibinfo{year}{2023}\natexlab{}.
\newblock \showarticletitle{Towards explainable and safe conversational agents for mental health: A survey}.
\newblock \bibinfo{journal}{\emph{arXiv preprint arXiv:2304.13191}} (\bibinfo{year}{2023}).
\newblock


\bibitem[Shao et~al\mbox{.}(2018)]%
        {shao2018find}
\bibfield{author}{\bibinfo{person}{Dian Shao}, \bibinfo{person}{Yu Xiong}, \bibinfo{person}{Yue Zhao}, \bibinfo{person}{Qingqiu Huang}, \bibinfo{person}{Yu Qiao}, {and} \bibinfo{person}{Dahua Lin}.} \bibinfo{year}{2018}\natexlab{}.
\newblock \showarticletitle{Find and focus: Retrieve and localize video events with natural language queries}. In \bibinfo{booktitle}{\emph{Proceedings of the European Conference on Computer Vision (ECCV)}}. \bibinfo{pages}{200--216}.
\newblock


\bibitem[Shao et~al\mbox{.}(2020a)]%
        {shao2020finegym}
\bibfield{author}{\bibinfo{person}{Dian Shao}, \bibinfo{person}{Yue Zhao}, \bibinfo{person}{Bo Dai}, {and} \bibinfo{person}{Dahua Lin}.} \bibinfo{year}{2020}\natexlab{a}.
\newblock \showarticletitle{Finegym: A hierarchical video dataset for fine-grained action understanding}. In \bibinfo{booktitle}{\emph{Proceedings of the IEEE/CVF conference on computer vision and pattern recognition}}. \bibinfo{pages}{2616--2625}.
\newblock


\bibitem[Shao et~al\mbox{.}(2020b)]%
        {shao2020intra}
\bibfield{author}{\bibinfo{person}{Dian Shao}, \bibinfo{person}{Yue Zhao}, \bibinfo{person}{Bo Dai}, {and} \bibinfo{person}{Dahua Lin}.} \bibinfo{year}{2020}\natexlab{b}.
\newblock \showarticletitle{Intra-and inter-action understanding via temporal action parsing}. In \bibinfo{booktitle}{\emph{Proceedings of the IEEE/CVF Conference on Computer Vision and Pattern Recognition}}. \bibinfo{pages}{730--739}.
\newblock


\bibitem[Simonyan and Zisserman(2014)]%
        {simonyan2014very}
\bibfield{author}{\bibinfo{person}{Karen Simonyan} {and} \bibinfo{person}{Andrew Zisserman}.} \bibinfo{year}{2014}\natexlab{}.
\newblock \showarticletitle{Very deep convolutional networks for large-scale image recognition}.
\newblock \bibinfo{journal}{\emph{arXiv preprint arXiv:1409.1556}} (\bibinfo{year}{2014}).
\newblock


\bibitem[Sun et~al\mbox{.}(2023)]%
        {sun2023mae}
\bibfield{author}{\bibinfo{person}{Licai Sun}, \bibinfo{person}{Zheng Lian}, \bibinfo{person}{Bin Liu}, {and} \bibinfo{person}{Jianhua Tao}.} \bibinfo{year}{2023}\natexlab{}.
\newblock \showarticletitle{Mae-dfer: Efficient masked autoencoder for self-supervised dynamic facial expression recognition}. In \bibinfo{booktitle}{\emph{Proceedings of the 31st ACM International Conference on Multimedia}}. \bibinfo{pages}{6110--6121}.
\newblock


\bibitem[Sun et~al\mbox{.}(2020)]%
        {sun2020multi}
\bibfield{author}{\bibinfo{person}{Licai Sun}, \bibinfo{person}{Zheng Lian}, \bibinfo{person}{Jianhua Tao}, \bibinfo{person}{Bin Liu}, {and} \bibinfo{person}{Mingyue Niu}.} \bibinfo{year}{2020}\natexlab{}.
\newblock \showarticletitle{Multi-modal continuous dimensional emotion recognition using recurrent neural network and self-attention mechanism}. In \bibinfo{booktitle}{\emph{Proceedings of the 1st international on multimodal sentiment analysis in real-life media challenge and workshop}}. \bibinfo{pages}{27--34}.
\newblock


\bibitem[Tao et~al\mbox{.}(2023)]%
        {10.1145/3581783.3611972}
\bibfield{author}{\bibinfo{person}{Zeng Tao}, \bibinfo{person}{Yan Wang}, \bibinfo{person}{Zhaoyu Chen}, \bibinfo{person}{Boyang Wang}, \bibinfo{person}{Shaoqi Yan}, \bibinfo{person}{Kaixun Jiang}, \bibinfo{person}{Shuyong Gao}, {and} \bibinfo{person}{Wenqiang Zhang}.} \bibinfo{year}{2023}\natexlab{}.
\newblock \showarticletitle{Freq-HD: An Interpretable Frequency-based High-Dynamics Affective Clip Selection Method for in-the-Wild Facial Expression Recognition in Videos}. In \bibinfo{booktitle}{\emph{Proceedings of the 31st ACM International Conference on Multimedia}} (<conf-loc>, <city>Ottawa ON</city>, <country>Canada</country>, </conf-loc>) \emph{(\bibinfo{series}{MM '23})}. \bibinfo{publisher}{Association for Computing Machinery}, \bibinfo{address}{New York, NY, USA}, \bibinfo{pages}{843–852}.
\newblock
\showISBNx{9798400701085}
\urldef\tempurl%
\url{https://doi.org/10.1145/3581783.3611972}
\showDOI{\tempurl}


\bibitem[Tao et~al\mbox{.}(2024)]%
        {tao2024a3ligndferpioneeringcomprehensivedynamic}
\bibfield{author}{\bibinfo{person}{Zeng Tao}, \bibinfo{person}{Yan Wang}, \bibinfo{person}{Junxiong Lin}, \bibinfo{person}{Haoran Wang}, \bibinfo{person}{Xinji Mai}, \bibinfo{person}{Jiawen Yu}, \bibinfo{person}{Xuan Tong}, \bibinfo{person}{Ziheng Zhou}, \bibinfo{person}{Shaoqi Yan}, \bibinfo{person}{Qing Zhao}, \bibinfo{person}{Liyuan Han}, {and} \bibinfo{person}{Wenqiang Zhang}.} \bibinfo{year}{2024}\natexlab{}.
\newblock \bibinfo{title}{A$^{3}$lign-DFER: Pioneering Comprehensive Dynamic Affective Alignment for Dynamic Facial Expression Recognition with CLIP}.
\newblock
\newblock
\showeprint[arxiv]{2403.04294}~[cs.CV]
\urldef\tempurl%
\url{https://arxiv.org/abs/2403.04294}
\showURL{%
\tempurl}


\bibitem[Tran et~al\mbox{.}(2015)]%
        {tran2015learning}
\bibfield{author}{\bibinfo{person}{Du Tran}, \bibinfo{person}{Lubomir Bourdev}, \bibinfo{person}{Rob Fergus}, \bibinfo{person}{Lorenzo Torresani}, {and} \bibinfo{person}{Manohar Paluri}.} \bibinfo{year}{2015}\natexlab{}.
\newblock \showarticletitle{Learning spatiotemporal features with 3d convolutional networks}. In \bibinfo{booktitle}{\emph{Proceedings of the IEEE international conference on computer vision}}. \bibinfo{pages}{4489--4497}.
\newblock


\bibitem[Tran et~al\mbox{.}(2018)]%
        {tran2018closer}
\bibfield{author}{\bibinfo{person}{Du Tran}, \bibinfo{person}{Heng Wang}, \bibinfo{person}{Lorenzo Torresani}, \bibinfo{person}{Jamie Ray}, \bibinfo{person}{Yann LeCun}, {and} \bibinfo{person}{Manohar Paluri}.} \bibinfo{year}{2018}\natexlab{}.
\newblock \showarticletitle{A closer look at spatiotemporal convolutions for action recognition}. In \bibinfo{booktitle}{\emph{Proceedings of the IEEE conference on Computer Vision and Pattern Recognition}}. \bibinfo{pages}{6450--6459}.
\newblock


\bibitem[Wang et~al\mbox{.}(2023c)]%
        {wang2023prototype}
\bibfield{author}{\bibinfo{person}{Binglu Wang}, \bibinfo{person}{Kang Yang}, \bibinfo{person}{Yongqiang Zhao}, \bibinfo{person}{Teng Long}, {and} \bibinfo{person}{Xuelong Li}.} \bibinfo{year}{2023}\natexlab{c}.
\newblock \showarticletitle{Prototype-based intent perception}.
\newblock \bibinfo{journal}{\emph{IEEE Transactions on Multimedia}}  \bibinfo{volume}{25} (\bibinfo{year}{2023}), \bibinfo{pages}{8308--8319}.
\newblock


\bibitem[Wang et~al\mbox{.}(2023b)]%
        {wang2023driver}
\bibfield{author}{\bibinfo{person}{Chen Wang}, \bibinfo{person}{Liang Shao}, \bibinfo{person}{Jun Liu}, {and} \bibinfo{person}{Jiawei Xiang}.} \bibinfo{year}{2023}\natexlab{b}.
\newblock \showarticletitle{Driver abnormal behavior detection system using two-stage object detection}.
\newblock  (\bibinfo{year}{2023}).
\newblock


\bibitem[Wang et~al\mbox{.}(2023a)]%
        {wang2023rethinking}
\bibfield{author}{\bibinfo{person}{Hanyang Wang}, \bibinfo{person}{Bo Li}, \bibinfo{person}{Shuang Wu}, \bibinfo{person}{Siyuan Shen}, \bibinfo{person}{Feng Liu}, \bibinfo{person}{Shouhong Ding}, {and} \bibinfo{person}{Aimin Zhou}.} \bibinfo{year}{2023}\natexlab{a}.
\newblock \showarticletitle{Rethinking the learning paradigm for dynamic facial expression recognition}. In \bibinfo{booktitle}{\emph{Proceedings of the IEEE/CVF Conference on Computer Vision and Pattern Recognition}}. \bibinfo{pages}{17958--17968}.
\newblock


\bibitem[Wang et~al\mbox{.}(2024)]%
        {wang2024m2}
\bibfield{author}{\bibinfo{person}{Mengmeng Wang}, \bibinfo{person}{Jiazheng Xing}, \bibinfo{person}{Boyuan Jiang}, \bibinfo{person}{Jun Chen}, \bibinfo{person}{Jianbiao Mei}, \bibinfo{person}{Xingxing Zuo}, \bibinfo{person}{Guang Dai}, \bibinfo{person}{Jingdong Wang}, {and} \bibinfo{person}{Yong Liu}.} \bibinfo{year}{2024}\natexlab{}.
\newblock \showarticletitle{M2-CLIP: A Multimodal, Multi-task Adapting Framework for Video Action Recognition}.
\newblock \bibinfo{journal}{\emph{arXiv preprint arXiv:2401.11649}} (\bibinfo{year}{2024}).
\newblock


\bibitem[Wang et~al\mbox{.}(2022a)]%
        {wang2022systematic}
\bibfield{author}{\bibinfo{person}{Yan Wang}, \bibinfo{person}{Wei Song}, \bibinfo{person}{Wei Tao}, \bibinfo{person}{Antonio Liotta}, \bibinfo{person}{Dawei Yang}, \bibinfo{person}{Xinlei Li}, \bibinfo{person}{Shuyong Gao}, \bibinfo{person}{Yixuan Sun}, \bibinfo{person}{Weifeng Ge}, \bibinfo{person}{Wei Zhang}, {et~al\mbox{.}}} \bibinfo{year}{2022}\natexlab{a}.
\newblock \showarticletitle{A systematic review on affective computing: Emotion models, databases, and recent advances}.
\newblock \bibinfo{journal}{\emph{Information Fusion}}  \bibinfo{volume}{83} (\bibinfo{year}{2022}), \bibinfo{pages}{19--52}.
\newblock


\bibitem[Wang et~al\mbox{.}(2022b)]%
        {wang2022ferv39k}
\bibfield{author}{\bibinfo{person}{Yan Wang}, \bibinfo{person}{Yixuan Sun}, \bibinfo{person}{Yiwen Huang}, \bibinfo{person}{Zhongying Liu}, \bibinfo{person}{Shuyong Gao}, \bibinfo{person}{Wei Zhang}, \bibinfo{person}{Weifeng Ge}, {and} \bibinfo{person}{Wenqiang Zhang}.} \bibinfo{year}{2022}\natexlab{b}.
\newblock \showarticletitle{Ferv39k: A large-scale multi-scene dataset for facial expression recognition in videos}. In \bibinfo{booktitle}{\emph{Proceedings of the IEEE/CVF conference on computer vision and pattern recognition}}. \bibinfo{pages}{20922--20931}.
\newblock


\bibitem[Wang et~al\mbox{.}(2022c)]%
        {10.1145/3503161.3547865}
\bibfield{author}{\bibinfo{person}{Yan Wang}, \bibinfo{person}{Yixuan Sun}, \bibinfo{person}{Wei Song}, \bibinfo{person}{Shuyong Gao}, \bibinfo{person}{Yiwen Huang}, \bibinfo{person}{Zhaoyu Chen}, \bibinfo{person}{Weifeng Ge}, {and} \bibinfo{person}{Wenqiang Zhang}.} \bibinfo{year}{2022}\natexlab{c}.
\newblock \showarticletitle{DPCNet: Dual Path Multi-Excitation Collaborative Network for Facial Expression Representation Learning in Videos}. In \bibinfo{booktitle}{\emph{Proceedings of the 30th ACM International Conference on Multimedia}} (<conf-loc>, <city>Lisboa</city>, <country>Portugal</country>, </conf-loc>) \emph{(\bibinfo{series}{MM '22})}. \bibinfo{publisher}{Association for Computing Machinery}, \bibinfo{address}{New York, NY, USA}, \bibinfo{pages}{101–110}.
\newblock
\showISBNx{9781450392037}
\urldef\tempurl%
\url{https://doi.org/10.1145/3503161.3547865}
\showDOI{\tempurl}


\bibitem[Yan et~al\mbox{.}(2024a)]%
        {yan2024empower}
\bibfield{author}{\bibinfo{person}{Shaoqi Yan}, \bibinfo{person}{Yan Wang}, \bibinfo{person}{Xinji Mai}, \bibinfo{person}{Qing Zhao}, \bibinfo{person}{Wei Song}, \bibinfo{person}{Jun Huang}, \bibinfo{person}{Zeng Tao}, \bibinfo{person}{Haoran Wang}, \bibinfo{person}{Shuyong Gao}, {and} \bibinfo{person}{Wenqiang Zhang}.} \bibinfo{year}{2024}\natexlab{a}.
\newblock \showarticletitle{Empower smart cities with sampling-wise dynamic facial expression recognition via frame-sequence contrastive learning}.
\newblock \bibinfo{journal}{\emph{Computer Communications}}  \bibinfo{volume}{216} (\bibinfo{year}{2024}), \bibinfo{pages}{130--139}.
\newblock


\bibitem[Yan et~al\mbox{.}(2024b)]%
        {yan2024urbanclip}
\bibfield{author}{\bibinfo{person}{Yibo Yan}, \bibinfo{person}{Haomin Wen}, \bibinfo{person}{Siru Zhong}, \bibinfo{person}{Wei Chen}, \bibinfo{person}{Haodong Chen}, \bibinfo{person}{Qingsong Wen}, \bibinfo{person}{Roger Zimmermann}, {and} \bibinfo{person}{Yuxuan Liang}.} \bibinfo{year}{2024}\natexlab{b}.
\newblock \showarticletitle{UrbanCLIP: Learning Text-enhanced Urban Region Profiling with Contrastive Language-Image Pretraining from the Web}. In \bibinfo{booktitle}{\emph{Proceedings of the ACM on Web Conference 2024}}. \bibinfo{pages}{4006--4017}.
\newblock


\bibitem[Yang et~al\mbox{.}(2023)]%
        {yang2023aim}
\bibfield{author}{\bibinfo{person}{Taojiannan Yang}, \bibinfo{person}{Yi Zhu}, \bibinfo{person}{Yusheng Xie}, \bibinfo{person}{Aston Zhang}, \bibinfo{person}{Chen Chen}, {and} \bibinfo{person}{Mu Li}.} \bibinfo{year}{2023}\natexlab{}.
\newblock \showarticletitle{Aim: Adapting image models for efficient video action recognition}.
\newblock \bibinfo{journal}{\emph{arXiv preprint arXiv:2302.03024}} (\bibinfo{year}{2023}).
\newblock


\bibitem[Yu et~al\mbox{.}(2024)]%
        {yu2024tf}
\bibfield{author}{\bibinfo{person}{Chenyang Yu}, \bibinfo{person}{Xuehu Liu}, \bibinfo{person}{Yingquan Wang}, \bibinfo{person}{Pingping Zhang}, {and} \bibinfo{person}{Huchuan Lu}.} \bibinfo{year}{2024}\natexlab{}.
\newblock \showarticletitle{TF-CLIP: Learning text-free CLIP for video-based person re-identification}. In \bibinfo{booktitle}{\emph{Proceedings of the AAAI Conference on Artificial Intelligence}}, Vol.~\bibinfo{volume}{38}. \bibinfo{pages}{6764--6772}.
\newblock


\bibitem[Zhang et~al\mbox{.}(2024b)]%
        {zhang2024two}
\bibfield{author}{\bibinfo{person}{Buyuan Zhang}, \bibinfo{person}{Haoyang Zhang}, \bibinfo{person}{Tao Zhen}, \bibinfo{person}{Bowen Ji}, \bibinfo{person}{Liang Xie}, \bibinfo{person}{Ye Yan}, {and} \bibinfo{person}{Erwei Yin}.} \bibinfo{year}{2024}\natexlab{b}.
\newblock \showarticletitle{A Two-Stage Real-Time Gesture Recognition Framework for UAV Control}.
\newblock \bibinfo{journal}{\emph{IEEE Sensors Journal}} (\bibinfo{year}{2024}).
\newblock


\bibitem[Zhang et~al\mbox{.}(2024a)]%
        {zhang2024long}
\bibfield{author}{\bibinfo{person}{Beichen Zhang}, \bibinfo{person}{Pan Zhang}, \bibinfo{person}{Xiaoyi Dong}, \bibinfo{person}{Yuhang Zang}, {and} \bibinfo{person}{Jiaqi Wang}.} \bibinfo{year}{2024}\natexlab{a}.
\newblock \showarticletitle{Long-clip: Unlocking the long-text capability of clip}.
\newblock \bibinfo{journal}{\emph{arXiv preprint arXiv:2403.15378}} (\bibinfo{year}{2024}).
\newblock


\bibitem[Zhang et~al\mbox{.}(2023)]%
        {damonlpsg2023videollama}
\bibfield{author}{\bibinfo{person}{Hang Zhang}, \bibinfo{person}{Xin Li}, {and} \bibinfo{person}{Lidong Bing}.} \bibinfo{year}{2023}\natexlab{}.
\newblock \showarticletitle{Video-LLaMA: An Instruction-tuned Audio-Visual Language Model for Video Understanding}.
\newblock \bibinfo{journal}{\emph{arXiv preprint arXiv:2306.02858}} (\bibinfo{year}{2023}).
\newblock
\urldef\tempurl%
\url{https://arxiv.org/abs/2306.02858}
\showURL{%
\tempurl}


\bibitem[Zhao and Liu(2021a)]%
        {zhao2021former}
\bibfield{author}{\bibinfo{person}{Zengqun Zhao} {and} \bibinfo{person}{Qingshan Liu}.} \bibinfo{year}{2021}\natexlab{a}.
\newblock \showarticletitle{Former-dfer: Dynamic facial expression recognition transformer}. In \bibinfo{booktitle}{\emph{Proceedings of the 29th ACM International Conference on Multimedia}}. \bibinfo{pages}{1553--1561}.
\newblock


\bibitem[Zhao and Liu(2021b)]%
        {10.1145/3474085.3475292}
\bibfield{author}{\bibinfo{person}{Zengqun Zhao} {and} \bibinfo{person}{Qingshan Liu}.} \bibinfo{year}{2021}\natexlab{b}.
\newblock \showarticletitle{Former-DFER: Dynamic Facial Expression Recognition Transformer}. In \bibinfo{booktitle}{\emph{Proceedings of the 29th ACM International Conference on Multimedia}} (Virtual Event, China) \emph{(\bibinfo{series}{MM '21})}. \bibinfo{publisher}{Association for Computing Machinery}, \bibinfo{address}{New York, NY, USA}, \bibinfo{pages}{1553–1561}.
\newblock
\showISBNx{9781450386517}
\urldef\tempurl%
\url{https://doi.org/10.1145/3474085.3475292}
\showDOI{\tempurl}


\bibitem[Zhao and Patras(2023)]%
        {zhao2023prompting}
\bibfield{author}{\bibinfo{person}{Zengqun Zhao} {and} \bibinfo{person}{Ioannis Patras}.} \bibinfo{year}{2023}\natexlab{}.
\newblock \showarticletitle{Prompting visual-language models for dynamic facial expression recognition}.
\newblock \bibinfo{journal}{\emph{arXiv preprint arXiv:2308.13382}} (\bibinfo{year}{2023}).
\newblock


\bibitem[Zheng et~al\mbox{.}(2022)]%
        {zheng2022general}
\bibfield{author}{\bibinfo{person}{Yinglin Zheng}, \bibinfo{person}{Hao Yang}, \bibinfo{person}{Ting Zhang}, \bibinfo{person}{Jianmin Bao}, \bibinfo{person}{Dongdong Chen}, \bibinfo{person}{Yangyu Huang}, \bibinfo{person}{Lu Yuan}, \bibinfo{person}{Dong Chen}, \bibinfo{person}{Ming Zeng}, {and} \bibinfo{person}{Fang Wen}.} \bibinfo{year}{2022}\natexlab{}.
\newblock \showarticletitle{General facial representation learning in a visual-linguistic manner}. In \bibinfo{booktitle}{\emph{Proceedings of the IEEE/CVF conference on computer vision and pattern recognition}}. \bibinfo{pages}{18697--18709}.
\newblock


\end{thebibliography}

\appendix
\newpage

\section{Appendix}
\subsection{Introduction}
The content of our supplementary material is organized as follows:

\noindent1) In Sec.~\ref{additional}, we present the variants illustration for FineCLIPER;

\noindent2) In Sec.~\ref{baseline}, we analyze two competitive baseline models and the efficiency of our FineCLIPER;

\noindent3) In Sec.~\ref{weight}, we further analyze the proposed adaptive weighting strategy and scaling factor $s$;

\noindent4) In Sec.~\ref{text}, we present detailed information regarding fine-grained text descriptions of facial action movements.

\subsection{FineCLIPER Variants}
\label{additional}


\textbf{FineCLIPER:} The variant utilizes only low-semantic level video frames along with PN (Postive-Negative) descriptors for label augmentation;

\noindent\textbf{FineCLIPER$^{\ast}$:} Building upon FineCLIPER, this variant incorporates middle-semantic level face parsing and landmarks;

\noindent\textbf{FineCLIPER$^{\dagger}$:} Extending FineCLIPER, this variant directly integrates high-semantic level fine-grained descriptions of facial action changes; 

\noindent\textbf{FineCLIPER$^{\ast}$$^{\dagger}$:} Expanding on FineCLIPER, this variant includes both middle-semantic level and high-semantic level information.

\subsection{Baselines vs. FineCLIPER} 
\label{baseline}
Here we first analyze two competitive baseline models:

\noindent \textbf{S2D}~\cite{chen2023static} achieved notable results with minimal tunable parameters on the ViT-B/16 backbone. However, it is noteworthy that it first undergoes pre-training on the Static Facial Expression Recognition (SFER) dataset, specifically AffectNet-7~\cite{Mollahosseini_2019} (consisting of 283,901 training samples) for 100 epochs, before fine-tuning on the DFER dataset. This pre-training step significantly contributes to its performance;

\noindent \textbf{A$^3$lign-DFER}~\cite{tao2024a3ligndferpioneeringcomprehensivedynamic}, as the latest CLIP-based DFER model, predominantly relies on the CLIP-ViT-L/14 backbone to further empower DFER from an alignment perspective. The training process is delineated into three stages spanning a total of 100 epochs. Regrettably, pertinent information regarding tunable parameters was not found within its paper.

\begin{figure}
    \centering
   \includegraphics[width=1\linewidth]{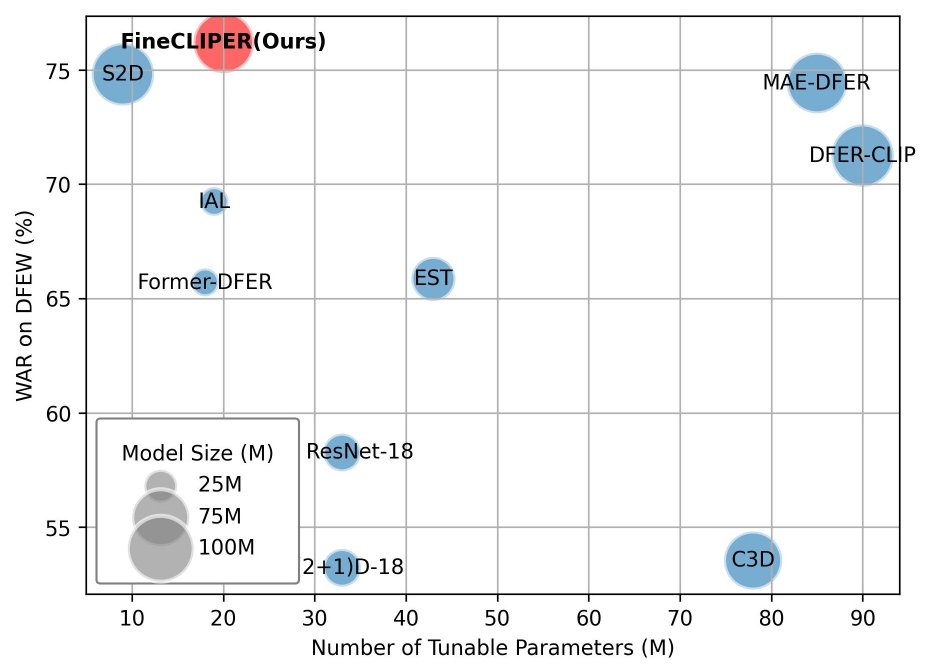}
    \captionof{figure}{\label{fig:para}
    Parameter-Performance comparison on the DFEW testing set. The bubble size indicates the model size.} 
    \vspace{-0.2cm}
\end{figure}

In contrast, our FineCLIPER model employs the CLIP-ViT-B/16 backbone and undergoes training solely on the DFER dataset for 30 epochs, achieving state-of-the-art performance with 13-20M tunable parameters in both supervised and zero-shot settings. Tab.~\ref{tab:app_model_size} presents the performance of FineCLIPER in the supervised setting on larger scales. Although a larger backbone could potentially yield better results, our choice prioritizes efficiency and ensures a fair comparison with baseline models. Additionally, we provide a parameter-performance comparison of DFER on the DFEW testing set in Fig.~\ref{fig:para}.

\subsection{Additional ablations}
\label{weight}
\textbf{Scaling factor $s$.} The scaling factor $s$ controls the weight of the output from the Adapter in Eq.~\ref{eq6}. In Tab.~\ref{tab:app_s}, we further conduct ablation experiments on the set value of $s$. More details can be found in~\cite{yang2023aim}.

\begin{figure}
    \centering
   \includegraphics[width=1\linewidth]{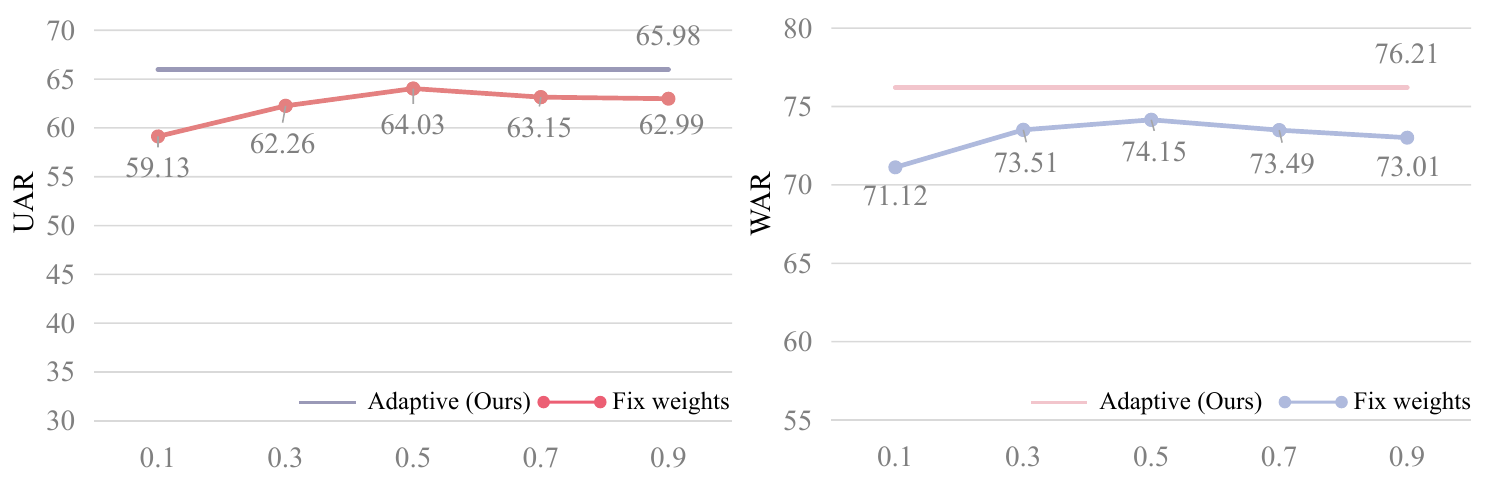}
    \captionof{figure}{\label{fig:weight}
    Comparison between our adaptive weighting strategy and fixed weights on the DFEW dataset, where the x-axis represents the weights of video features.} 
\end{figure}

\noindent\textbf{Adaptive Weighting.} Due to the inherent correlation between the expanded multimodal data, \textit{i.e.}, face parsing, landmarks, and fine-grained text, with videos, this work endeavors to explore the feasibility of further modeling faces using multi-modal data. In our proposed adaptive weighting algorithm, we determine the fusion of features and the weighting of the loss function adaptively by computing the similarity between multi-modal features and label features. To further validate the superiority of this strategy, we fix the weight of the video feature, $\mathbf{w_v}$, ranging from 0.1 to 0.9, while evenly distributing the remaining weight among the other three modal features, \textit{i.e.}, $1-\mathbf{w_v}$. 
As illustrated in Fig.~\ref{fig:weight}, our proposed adaptive weighting strategy exhibits greater stability and effectiveness compared to fixed weights.

Certainly, the fusion of features can be further optimized. The simplicity of our fusion strategy in this study serves to further substantiate the feasibility and potential of leveraging multi-modal data for DFER. In future works, we intend to delve deeper into the potential of feature fusion.

\begingroup
\setlength{\tabcolsep}{3pt}
\begin{table}
\renewcommand{\arraystretch}{1}
  \centering
  \caption{Ablation on model size.}
  \centering
  \small
   \begin{tabular}{lccccccc}
     \hlineB{2}
      \multirow{2}{*}{Backbone} & Tunable & \multicolumn{2}{c}{DFEW}& \multicolumn{2}{c}{FERV39k} & \multicolumn{2}{c}{MAFW}\\
     \cline{3-8}
     & Param (M) & UAR & WAR & UAR & WAR & UAR & WAR \\
     \hlineB{1.5}
     \rowcolor{cyan!10}
     CLIP-ViT-B/16 & 20 & 65.98 & 76.21 & 45.22 & 53.98 & 45.01 & 56.91 \\
     \rowcolor{cyan!10}
     CLIP-ViT-B/32 & 20 & 66.78 & 77.39 & 46.99 & 55.03 & 46.79 & 57.89 \\
     \rowcolor{cyan!10}
     CLIP-ViT-L/14 & 20 & \textbf{68.01} & \textbf{78.93} & \textbf{47.89} & \textbf{56.41} & \textbf{47.79} & \textbf{58.26} \\
     \hlineB{2}
   \end{tabular}
   \vspace{-0.2cm}
  \label{tab:app_model_size}
\end{table} 
\endgroup

\begingroup
\setlength{\tabcolsep}{8.2pt}
\begin{table}
\renewcommand{\arraystretch}{1}
  \centering
  \caption{Ablation on scaling factor $s$.}
  \centering
  \small
\begin{tabular}{lcccccc}
\hlineB{2}
\multirow{2}{*}{$s$} & \multicolumn{2}{c}{DFEW} & \multicolumn{2}{c}{FERV39k} & \multicolumn{2}{c}{MAFW} \\
\cline{2-7}
& UAR & WAR & UAR & WAR & UAR & WAR \\
\hlineB{1.5}
\rowcolor{yellow!10}
0.3 & 62.51 & 72.68 & 42.61 & 51.19 & 41.84 & 52.71 \\
\rowcolor{cyan!10}
0.5 & \textbf{62.81} & \textbf{72.86} & \textbf{42.88} & \textbf{52.01} & \textbf{42.19} & \textbf{53.12} \\
\rowcolor{yellow!10}
0.7 & 62.58 & 72.70 & 42.57 & 51.46 & 41.76 & 53.66 \\
\hlineB{2}
\end{tabular}
  \label{tab:app_s}
\end{table} 
\endgroup

\begin{figure}
    \centering
   \includegraphics[width=1\linewidth]{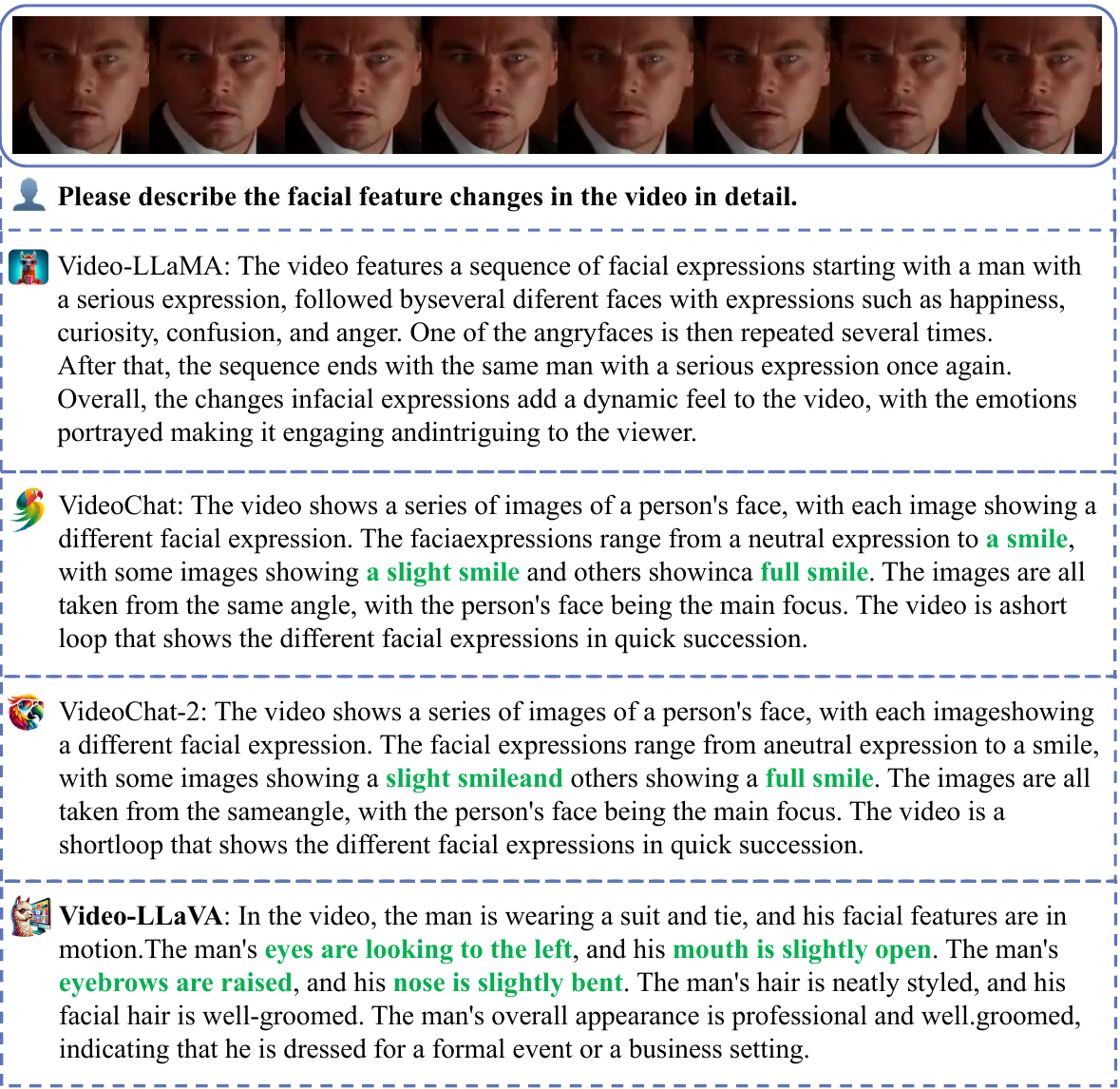}
    \captionof{figure}{\label{fig:mllmtext}
    Comparison of MLLM-generated captions for video. Facial expressions related are highlighted in \textcolor{green}{green}.} 
\end{figure}

\begin{figure}
    \centering
   \includegraphics[width=1\linewidth]{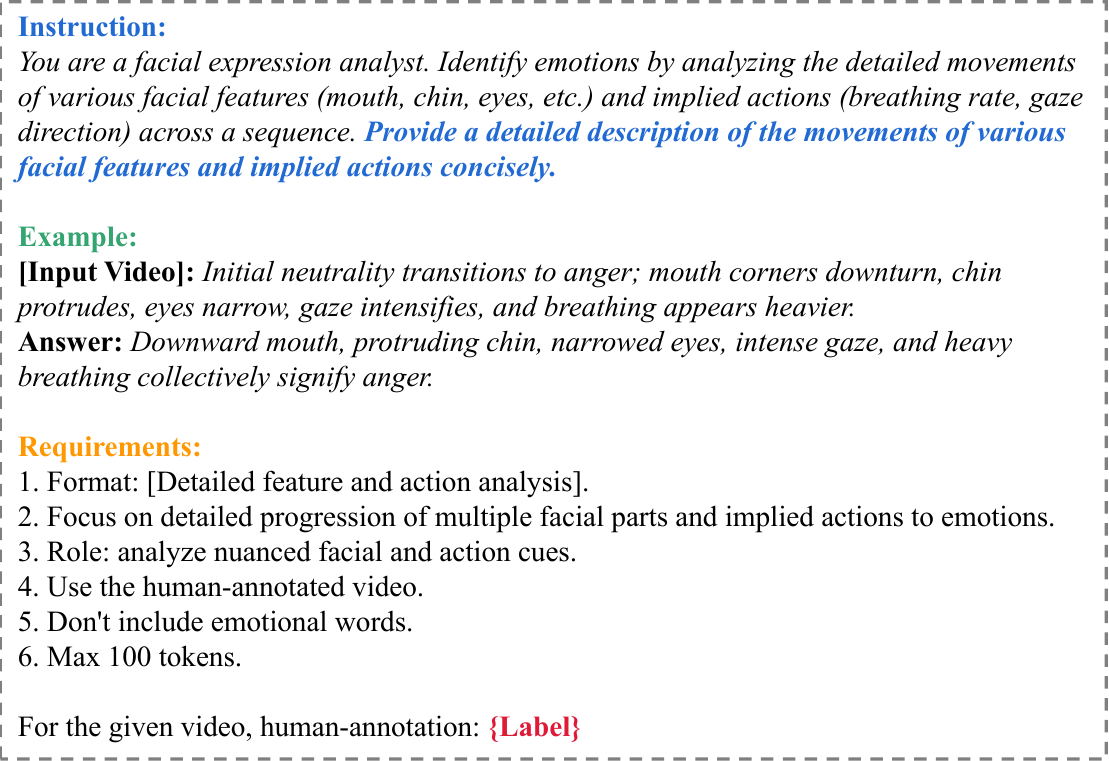}
    \captionof{figure}{\label{fig:prompt}
    Text Prompt Demonstration} 
\end{figure}

\begin{figure*}
    \centering
   \includegraphics[width=1\linewidth]{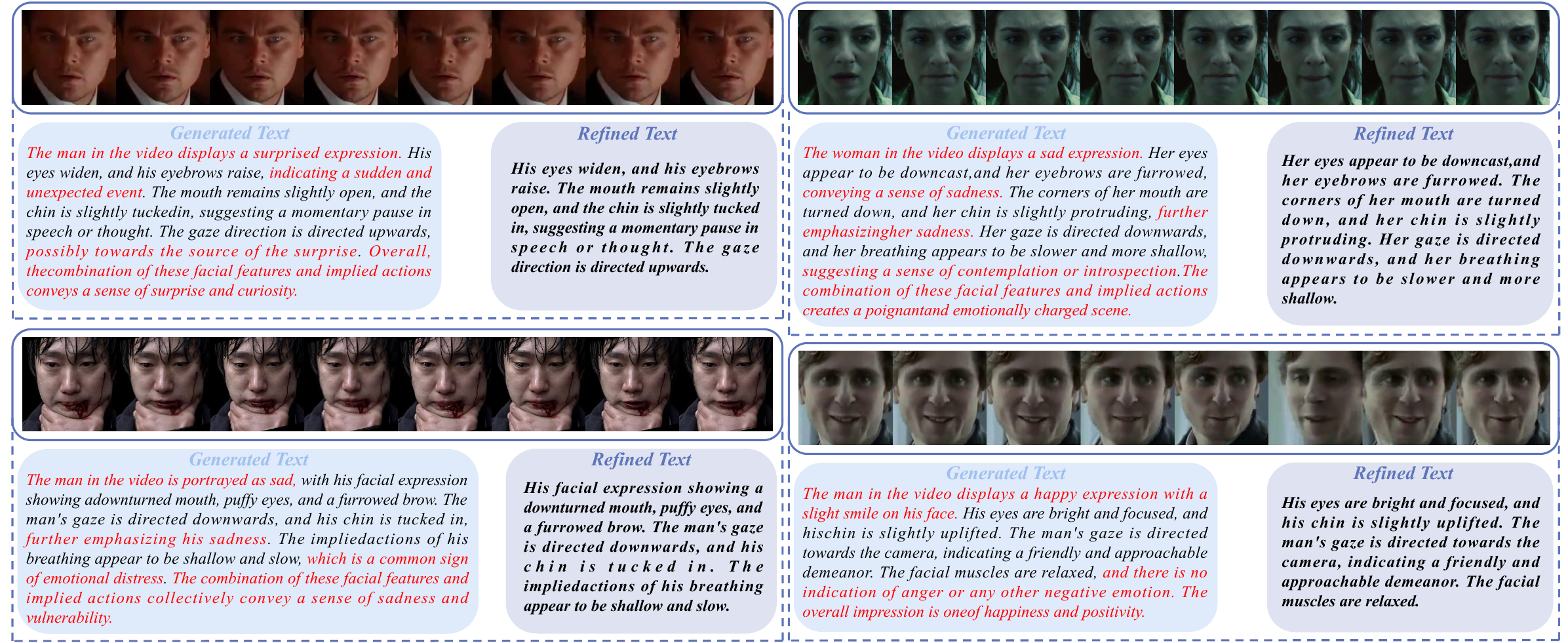}
    \captionof{figure}{\label{fig:exp}
    Examples of the generated text and the refined text.} 
\end{figure*}

\subsection{Fine-grained Text Generation}
\label{text}
\noindent \textbf{MLLM Selection.} 
Considering resource consumption, we initially evaluated several open-source Multi-modal Large Language Models (MLLMs) capable of processing videos, namely Video-LLaMA~\cite{damonlpsg2023videollama}, VideoChat~\cite{2023videochat}, VideoChat-2~\cite{li2024mvbench}, and Video-LLaVA~\cite{lin2023video}. To expedite the assessment of their ability to comprehend facial videos, we employed a simple prompt at this stage, \textit{i.e.}, "Please describe the facial feature changes in the video in detail". 
As depicted in Fig.~\ref{fig:mllmtext}, we highlight in green the descriptions of facial features outputted by the four MLLMs. It is evident that Video-LLaVA, compared to the other three, more accurately captures facial feature information. Consequently, we adopt Video-LLaVA as our fine-grained text generation model. Subsequent sections will elaborate on the detailed text prompt and refinement process for fine-grained text generation.

\noindent \textbf{Prompt Design.}
With the advancement of large language models, the significance of prompt engineering has become increasingly apparent. Well-crafted prompts can significantly enhance a model's ability to generate responses tailored to specific tasks. As illustrated in Fig.~\ref{fig:prompt}, in addition to explicit instructions, corresponding examples are provided for the model to reference and learn from. Furthermore, to further standardize the model's responses, six requirements are delineated. Finally, considering that the model may describe actions based on its analysis of facial expressions, ground truth labels for each video are also provided for the model's reference.

\noindent \textbf{Text Refinement.} 
Text refinement plays a pivotal role in our proposed FineCLIPER framework. Specifically, we identify two categories of low-quality text: 
1) Directly expressing emotions. For example, stating "The man in the video wears a sad expression..." This can lead to data leakage during the training process.
2) Indirectly implying emotions. For example, stating "The man's mouth is slightly ajar, showing his teeth, and his eyes are narrowed, suggesting a feeling of joy or amusement." Despite not explicitly containing label information, such descriptions still pose a risk of potential data leakage.

To this end, we introduce a two-stage heuristic process for text refinement in this study, as outlined by~\cite{yan2024urbanclip}, which comprises text cleaning and counterfactual verification as illustrated in the main content. Specifically, manual inspection involves the participation of numerous master's and undergraduate students with backgrounds in psychology or computer science.

\noindent \textbf{Generated Instances.}
As demonstrated in Fig.\ref{fig:exp}, despite the careful design of the prompt in Fig.\ref{fig:prompt}, emphasizing "Don't include emotional words," the generated text still contains several direct or indirect emotional expressions, as highlighted in \textcolor{red}{red}. Subsequently, through the implementation of the two-stage text refinement process, the refined text predominantly encompasses facial features and implied actions, as highlighted in \textbf{bold}. Such refined fine-grained text significantly enhances and strengthens facial modeling from a high semantic level. All the fine-grained descriptions, along with the face parsing and landmarks data will be released after the paper notification.









\end{document}